\documentclass[manuscript,screen]{acmart}

\AtBeginDocument{%
  \providecommand\BibTeX{{%
    \normalfont B\kern-0.5em{\scshape i\kern-0.25em b}\kern-0.8em\TeX}}}

\setcopyright{acmcopyright}
\copyrightyear{2024}
\acmYear{2024}
\acmDOI{XXXXXXX.XXXXXXX}

\usepackage{bm,multirow,tablefootnote}
\usepackage{longtable}
\usepackage{bbding}
\usepackage{threeparttable} 
\usepackage[para]{footmisc}
\usepackage{geometry}
\usepackage{makecell}

\begin{document}

\title{Generalized Video Anomaly Event Detection: Systematic Taxonomy and Comparison of Deep Models}

\author{Yang Liu}
\authornote{This paper was revised by Yang Liu during his FDU-UofT Joint Ph.D. Training Program at the University of Toronto, Canada.}
\email{yang\_liu20@fudan.edu.cn}
\orcid{0000-0002-1312-0146}
\affiliation{%
  \institution{Fudan University}
  \city{Shanghai}
  \postcode{200433}
  \country{China}
}
\affiliation{%
  \institution{University of Toronto}
  \city{Toronto}
  \postcode{M5S 1A1}
  \state{Ontario}
  \country{Canada}
}

\author{Dingkang Yang}
\email{dkyang20@fudan.edu.cn}
\orcid{0000-0003-1829-5671}
\author{Yan Wang}
\email{yanwang19@fudan.edu.cn}
\orcid{0000-0002-4953-2660}
\affiliation{%
  \institution{Fudan University}
  \city{Shanghai}
  \postcode{200433}
  \country{China}
}

\author{Jing Liu}
\email{jingliu19@fudan.edu.cn}
\orcid{0000-0002-2819-0200}
\affiliation{%
  \institution{Fudan University}
  \city{Shanghai}
  \postcode{200433}
  \country{China}
}
\affiliation{%
  \institution{University of British Columbia}
  \city{Vancouver}
  \postcode{V6T 1Z4}
  \state{British Columbia}
  \country{Canada}
}
\affiliation{%
  \institution{Duke Kunshan University}
  \city{Suzhou}
  \postcode{215316}
  \state{Jiangsu}
  \country{China}
}

\author{Jun Liu}
\email{jun\_liu@sutd.edu.sg}
\orcid{0000-0002-4365-4165}
\affiliation{%
  \institution{Singapore University of Technology and Design}
  \city{Singapore}
  \postcode{487372}
  \country{Singapore}
}

\author{Azzedine Boukerche}
\email{aboukerc@uOttawa.ca}
\orcid{0000-0002-3851-9938}
\affiliation{%
  \institution{University of Ottawa}
  \city{Ottawa}
  \postcode{K1N 6N5}
  \state{Ontario}
  \country{Canada}
}

\author{Peng Sun}
\authornote{Prof. Peng Sun and Prof. Liang Song are the co-corresponding authors of this paper.}
\email{peng.sun568@duke.edu}
\orcid{0000-0001-8356-1329}
\affiliation{%
  \institution{Duke Kunshan University}
  \city{Suzhou}
  \postcode{215316}
  \state{Jiangsu}
  \country{China}
}

\author{Liang Song}
\authornotemark[2]
\email{songl@fudan.edu.cn}
\orcid{0000-0002-8143-9052}
\affiliation{%
  \institution{Fudan University}
  \city{Shanghai}
  \postcode{200433}
  \country{China}
}

\renewcommand{\shortauthors}{Yang Liu \textit{ et al.}}
\linepenalty=1000

\begin{abstract}
  Video Anomaly Detection (VAD) serves as a pivotal technology in the intelligent surveillance systems, enabling the temporal or spatial identification of anomalous events within videos. While existing reviews predominantly concentrate on conventional unsupervised methods, they often overlook the emergence of weakly-supervised and fully-unsupervised approaches. To address this gap, this survey extends the conventional scope of VAD beyond unsupervised methods, encompassing a broader spectrum termed Generalized Video Anomaly Event Detection (GVAED). By skillfully incorporating recent advancements rooted in diverse assumptions and learning frameworks, this survey introduces an intuitive taxonomy that seamlessly navigates through unsupervised, weakly-supervised, supervised and fully-unsupervised VAD methodologies, elucidating the distinctions and interconnections within these research trajectories. In addition, this survey facilitates prospective researchers by assembling a compilation of research resources, including public datasets, available codebases, programming tools, and pertinent literature. Furthermore, this survey quantitatively assesses model performance, delves into research challenges and directions, and outlines potential avenues for future exploration.
\end{abstract}




\maketitle

\section{Introduction}
Surveillance cameras can sense environmental spatial-temporal information without contact and have been the primary data collection tool for public services such as security protection \cite{UMN}, crime warning \cite{MIR}, and traffic management \cite{STM-AE}. However, with the rapid development of smart cities and digital society, the number of surveillance cameras is growing explosively, making the ensuing video analysis a significant challenge. Traditional manual inspection is time-consuming and laborious and may cause missing detections due to human visual fatigue \cite{liu2022collaborative}, hardly coping with the vast scale video stream. As the core technology of intelligent surveillance systems, Video Anomaly Detection (VAD) aims to automatically analyze video patterns and locate abnormal events. Due to its potential application in unmanned factories, self-driving vehicles, and secure communities, VAD has received wide attention from academia and industry. 

VAD in a narrow sense refers specifically to the unsupervised research paradigm that uses only normal videos to learn a normality model, abbreviated as UVAD. Such methods share the same assumption as the long-established Anomaly Detection (AD) tasks in non-visual data (e.g., time series \cite{blazquez2021review} and graphs \cite{ma2021comprehensive}) and images \cite{huang2021self}. They assume the normality model learned on normal samples cannot represent anomalous samples. Typically, UVAD consists of two phases, normality learning and downstream anomaly detection \cite{FF-AE,FFP,STM-AE,yang2023video}. UAVD shares a similar modeling process with other AD tasks without predefining and collecting anomalies, following the open-world principle. In the real world, anomalies are diverse and rare, so they cannot be defined and fully collected in advance. Therefore, UVAD was favored by early researchers and was once considered the prevailing VAD paradigm. However, the definition of anomaly is idealistic, ignoring that normal events are diverse. It is also unrealistic to collect all possible regular events for modeling. In addition, the learned UVAD model has difficulty maintaining a reasonable balance between representation and generalization power, either due to the insufficient representational that false-alarms unseen normal events as anomalies or the excessive generalization power that effectively reconstructs anomalous events. Numerous experiments \cite{STC-Net} have shown that UVAD is valid for only simple scenarios. The model performance on complex datasets \cite{FFP} is much inferior to that of simple single-scene videos \cite{T2,T3}, which limits the application of VAD technicals in realistic scenarios.

In contrast, Weakly-supervised Abnormal Event Detection (WAED) departs from the ambiguous setting that all are anomalous except normal with a clearer definition for the anomaly that is more consistent with human consciousness (e.g., traffic accidents, robbery, stealing, and shooting). \cite{MIR}. Given its potential for immediate references in real-life applications such as traffic management platforms and violence warning systems, WAED has become another mainstream VAD paradigm since 2018 \cite{MIST,RTFM,SMR}. Generally, WAED models directly output anomaly scores by comparing the spatial-temporal features of normal and abnormal events through Multiple Instance Learning (MIL). The previous study \cite{MIST} proved that WAED could understand the essential difference between normal and abnormal. Therefore, its results are more reliable than that of UVAD. Unfortunately, WAED does not follow the basic assumptions of AD tasks, which is more like a binary classification under unique settings (e.g., data imbalance and positive samples containing multiple subcategories). Therefore, existing reviews \cite{ramachandra2020survey,chandrakala2022anomaly,raja2022analysis} mainly focus on UVAD and consider WAED as a marginal research pathway, lacking the systematic organization for WAED datasets and methods. 

\begin{figure}[t]
  \centering
  \includegraphics[width=\textwidth]{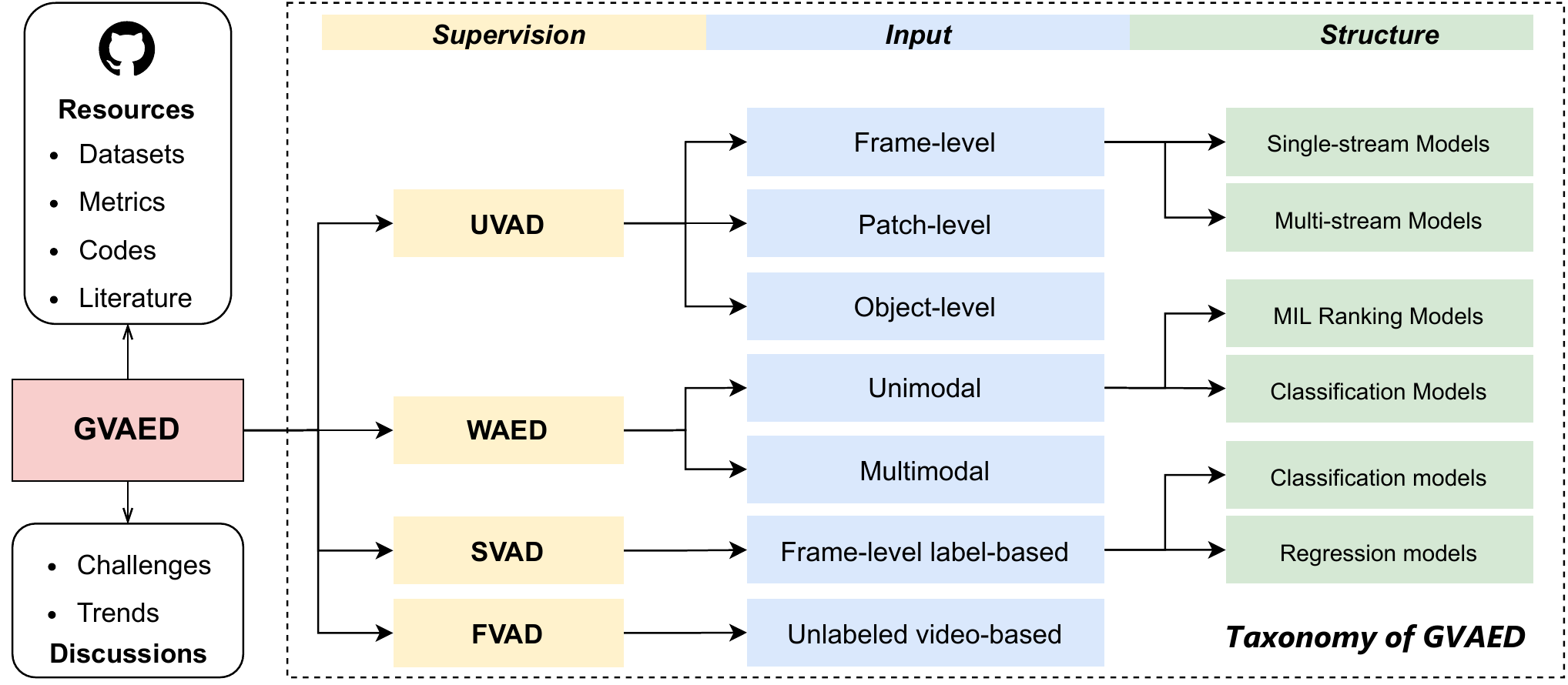}
  \caption{Taxonomy of Generalized Video Anomaly Event Detection (GVAED). We provide a hierarchical taxonomy that organizes existing deep GVAED models by supervised signals, model inputs, and network structure into a systematic framework, including Unsupervised Video Anomaly Detection (UVAD), Weakly-supervised Abnormal Event Detection (WAED), Fully-unsupervised VAD (FVAD) and Supervised VAD (SVAD). Besides, we collate benchmark datasets, evaluation metrics, available codes, and literature to a public GitHub repository$^1$. Finally, we analyze the research challenges and possible trends.}
  \label{fig:overview}
\end{figure}

In recent times, certain researchers \cite{GCL,SDOR} have introduced fully-unsupervised methods, which eliminate the need for labels and eliminate any prerequisites on the training data. Given the deeply ingrained association of UVAD with modeling using solely normal data, we retain the nomenclature UVAD for these techniques, rather than referring to them as semi-supervised VAD. Conversely, the emerging paradigm of absolute unsupervised setting is denoted as FVAD, contributing to terminology conceptual clarity in this survey and future research.

In summary, this survey focuses on anomaly event detection in surveillance videos, integrating deep VAD methods based on different assumptions, learning paradigms, and supervision into a systematic taxonomy: Generalized Video Anomaly Event Detection (GVAED), as shown in Fig. \ref{fig:overview}. We compare the differences and performance among different methods, sorting out the recent advances in GVAED. In addition, we collate available research resources, such as datasets, metrics, codes, and literature, into a public GitHub repository\footnote[1]{\url{https://github.com/fudanyliu/GVAED.git}}. Moreover, we analyze the research challenges and future trends, which can guide further research and promote the development and applications of GVAED.

\subsection{Literature Statistics}

\begin{figure}
  \centering
  \includegraphics[width=\textwidth]{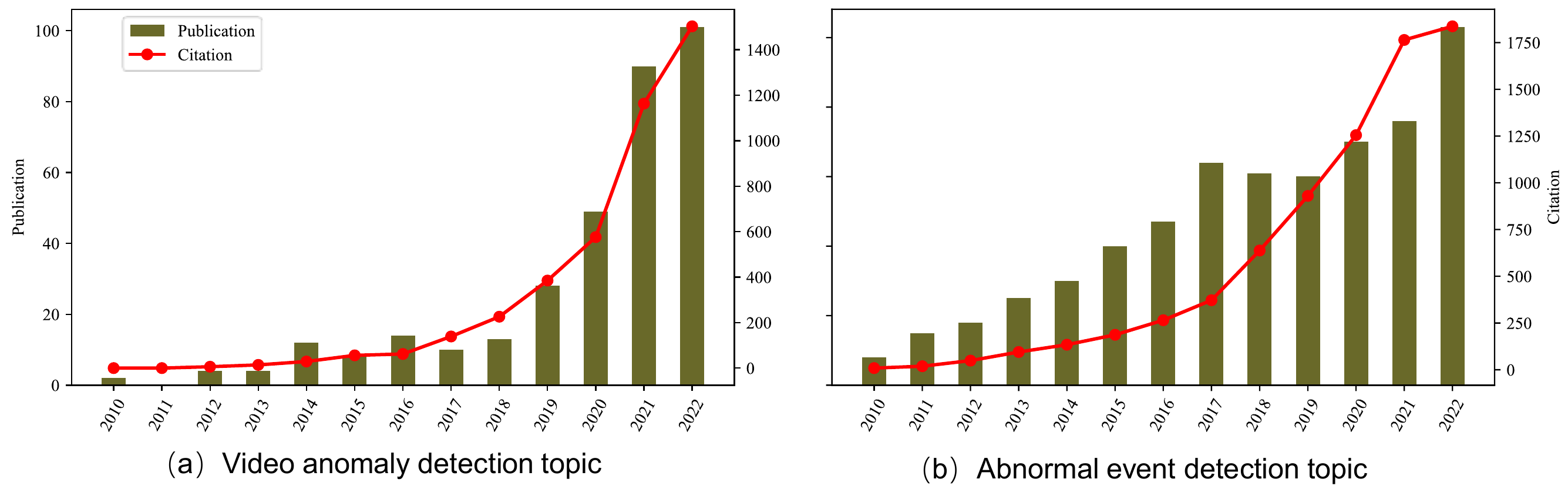}
  \caption{Publication and citation statistics on the topic of (a) \textit{Video Anomaly Detection} and (b) \textit{Abnormal Event Detection}.}
  \label{fig:wxtj}
\end{figure}

We count the publications and citations of academic papers on the topic of \textit{Video Anomaly Detection} and \textit{Abnormal Event Detection} in the past 12 years through reference databases (e.g., ACM Digital Library, IEEE explore, ScienceDirect, and SpringerLink) and search engines. The results are shown in Fig.~\ref{fig:wxtj}, where the bar and line graph indicate the number of publications and citations, respectively. The lines in Fig.~\ref{fig:wxtj}(a) and Fig.~\ref{fig:wxtj}(b) show a steadily increasing trend, indicating that GVAED has received wide attention. Therefore, a systematic taxonomy and comparison of GVAED methods are necessary to guide further development. As mentioned above, current works focus on unsupervised methods that use only regular videos to train models to represent normality. Thus, the development of UVAD is limited by representation means. Until 2016, UVAD utilized handicraft features, such as Local Binary Patterns (LBP) \cite{hu2018squirrel,nawarathna2014abnormal,zhang2015efficient}, Histogram Of Gradients (HOG) \cite{FF-AE,LGA,T38}, and Space-Time Interest Point (STIP) \cite{T72}. The performance is poor and relies on a priori knowledge. As a result, VAD developed slowly. Fortunately, after 2016, with the development of deep learning, especially the application of Convolutional Neural Networks (CNNs) in image processing \cite{chen2022towards,li2022adaptive,yang2022emotion} and video understanding \cite{yang2023target,yang2023spatio,yang2023aide}, VAD has ushered in new development opportunities. The research progress increased significantly, as shown in Fig~\ref{fig:wxtj}(a). Deep CNNs can extract the video patterns end-to-end, freeing VAD research from complex a priori knowledge construction. In addition, compared with manual features \cite{hu2018squirrel,T72,LGA}, deep representations can capture multi-scale spatial semantic features and extract long-range temporal contextual features, which are more efficient in learning video normality. On the one hand, the large amount of video generated by the surveillance cameras provides sufficient training data for deep GVAED models. On the other hand, the iteratively updated Graphics Processing Units (GPUs) make it possible to train large-scale models. As a result, VAD has developed rapidly in recent years and started to move from academic research to commercial applications. Similarly, Fig~\ref{fig:wxtj}(b) reflects the research enthusiasm and development potential of abnormal event detection. 
To accelerate the application of GVAED in terminal devices and inspire future researchers, this survey organizes various GVAED models into a unified framework. Additionally, we collect commonly used datasets, publicly available codes, and classic literature for further research.

\subsection{Related Reviews}
\begin{table}[]
  \centering
  \caption{Analysis and Comparison of Related Reviews.}
  \label{tab:xgzs}
\begin{threeparttable}
  \resizebox{\textwidth}{!}{
  \begin{tabular}{cllcccc|cccc}
  \hline
  \multirow{2}{*}{\textbf{Year}} & \multirow{2}{*}{\textbf{Ref.}}        & \multicolumn{1}{c}{\multirow{2}{*}{\textbf{Main Focus}}} & \multicolumn{4}{c|}{\textbf{Research Pathways}$^a$}         & \multicolumn{4}{c}{\textbf{Topics}$^{a,b}$} \\ \cline{4-11} 
      &   & \multicolumn{1}{l}{}    & UVAD      & WAED      & SVAD      & FVAD      & LW      & OD      & CS     & OE     \\ \hline
  2018         & \cite{kiran2018overview}      & Unsupervised and semi-supervised methods.         & \footnotesize\Checkmark   & \footnotesize\XSolidBrush & \footnotesize\XSolidBrush & \footnotesize\XSolidBrush &  \footnotesize\Checkmark       &   \footnotesize\XSolidBrush      &   \footnotesize\XSolidBrush     &   \footnotesize\XSolidBrush     \\
  2019         & \cite{lomte2019survey}        & Weakly-supervised VAD methods and applications.   & \footnotesize\XSolidBrush & \footnotesize\Checkmark   & \footnotesize\XSolidBrush & \footnotesize\XSolidBrush &    \footnotesize\XSolidBrush     & \footnotesize\XSolidBrush        &  \footnotesize\XSolidBrush      &  \footnotesize\XSolidBrush     \\
  2020         & \cite{ramachandra2020survey}  & Unsupervised single-scene video anomaly detection.         & \footnotesize\Checkmark   & \footnotesize\XSolidBrush & \footnotesize\Checkmark   & \footnotesize\XSolidBrush &  \footnotesize\Checkmark       &    \footnotesize\XSolidBrush     &   \footnotesize\Checkmark     &    \footnotesize\XSolidBrush    \\
  2021         & \cite{nayak2021comprehensive} & Deep learning driven unsupervised VAD methods.   & \footnotesize\Checkmark   & \footnotesize\XSolidBrush & \footnotesize\XSolidBrush & \footnotesize\XSolidBrush &  \footnotesize\Checkmark       &   \footnotesize\XSolidBrush      &     \footnotesize\XSolidBrush   &    \footnotesize\XSolidBrush    \\
  2021         & \cite{rezaee2021survey}       & Unsupervised crowd anomaly detection methods.  & \footnotesize\Checkmark   & \footnotesize\XSolidBrush & \footnotesize\XSolidBrush & \footnotesize\XSolidBrush &     \footnotesize\XSolidBrush    &   \footnotesize\XSolidBrush      &  \footnotesize\XSolidBrush      &    \footnotesize\XSolidBrush    \\
  2021         & \cite{santhosh2020anomaly}    & Traffic scene video anomaly detection.  & \footnotesize\Checkmark   & \footnotesize\XSolidBrush & \footnotesize\XSolidBrush & \footnotesize\XSolidBrush &     \footnotesize\XSolidBrush    &    \footnotesize\XSolidBrush     &   \footnotesize\XSolidBrush     &   \footnotesize\XSolidBrush     \\
  2022         & \cite{raja2022analysis}       & Unsupervised video anomaly detection.    & \footnotesize\Checkmark   & \footnotesize\XSolidBrush & \footnotesize\XSolidBrush & \footnotesize\XSolidBrush &    \footnotesize\XSolidBrush     &  \footnotesize\Checkmark       &     \footnotesize\XSolidBrush   &     \footnotesize\XSolidBrush   \\
  2022         & \cite{chandrakala2022anomaly} & One\&two-class classification-based methods.    & \footnotesize\Checkmark   & \footnotesize\Checkmark   & \footnotesize\Checkmark   & \footnotesize\XSolidBrush &   \footnotesize\Checkmark      &   \footnotesize\Checkmark      &      \footnotesize\XSolidBrush  &     \footnotesize\XSolidBrush   \\ \hline
  2023         & Ours   & GVAED taxonomy, challenges and trends.   & \footnotesize\Checkmark   & \footnotesize\Checkmark   & \footnotesize\Checkmark   & \footnotesize\Checkmark   &  \footnotesize\Checkmark       &   \footnotesize\Checkmark      &   \footnotesize\Checkmark     &  \footnotesize\Checkmark      \\ \hline
  \end{tabular}}
  \begin{tablenotes}    
    \footnotesize   
    \item a: \footnotesize\XSolidBrush means no systematic analysis, while \footnotesize\Checkmark is vice versa. b: LW=lightweight, OD=online detection, CS=cross-scene, and OE=online evolution.
  \end{tablenotes}   
\end{threeparttable}
\end{table}

In the past four years, several reviews \cite{kiran2018overview,cook2019anomaly,nayak2021comprehensive,rezaee2021survey,santhosh2020anomaly,pang2021deep,blazquez2021review,ramachandra2020survey,chandrakala2022anomaly,raja2022analysis,modi2022survey,jebur2022review,sharif2022deep,lomte2019survey} have covered GVAED works and generated various classification systems. We analyze the methodologies covered in recent reviews and the research topics related to real-world deployment, as shown in Table~\ref{tab:xgzs}. The mainstream reviews \cite{kiran2018overview,ramachandra2020survey} still consider VAD as a narrow unsupervised task, lacking attention to WAED with excellent application value and FVAD methods using unlimited training data. In addition, they are biased against Supervised Video Anomaly Detection (SVAD), arguing that data labeling makes SVAD challenging to develop. However, the game engines \cite{engine1,engine2} and automatic annotations \cite{auto1,auto2} make it possible to obtain anomalous events and fine-grained labels. In addition, the existing review suffers from three major weaknesses: (1) \cite{ramachandra2020survey} and \cite{santhosh2020anomaly} attempted to link existing works to specific scenes, missing the cross-scene challenges in real-world. Specifically, \cite{ramachandra2020survey} pointed out that existing works were trained and tested on videos of the same scene, so they only reviewed single-single methods, leaving out the latest cross-scene VAD research. \cite{santhosh2020anomaly} focuses on the traffic VAD methods, innovatively analyzing the applicability of existing works in traffic scenes. However, weakly-supervised methods for crime and violence detection fail to be included in \cite{santhosh2020anomaly}. (2) Due to timeliness, earlier reviews \cite{cook2019anomaly,nayak2021comprehensive,rezaee2021survey,santhosh2020anomaly} were unable to cover the latest research and were outdated for predicting research trends. Recent surveys \cite{chandrakala2022anomaly,raja2022analysis} lack discussion of the interaction of GVAED with new techniques such as causal machine learning \cite{aaai21,liu2023learning}, domain adaptation \cite{Background-Agnostic,wilson2020survey}, and online evolutive learning \cite{li2022multi,li2022mutual,song2022networking}, which are expected to be the future directions of GVAED and essential to model deployment. (3) Although the latest review \cite{chandrakala2022anomaly} in 2022 has started to incorporate WAED and SVAD, it still treats them as a marginal exploration, lacking a systematic organization of the datasets, literature, and trends. 

\subsection{Contribution Summary}
GVAED will usher in new development opportunities with the rapid growth of deep learning technicals and surveillance videos. 
To clarify the development of GVAED and inspire future research, this survey integrates UVAD, WAED, FVAD, and SVAD into a unified framework from an application perspective. The main contributions of this survey are summarized in the following four aspects:
\begin{enumerate}
  \item To the best of our knowledge, it is the first comprehensive survey that extends video anomaly detection from narrow unsupervised methods to generalized video anomaly event detection. We analyze the various research routes and clearly state the lineage and trends of deep GVAED models to help advance the field.
  \item We organize various GVAED methods with different assumptions and learning frameworks from an application perspective, providing an intuitive taxonomy based on supervision signals, input data, and network structures.
  \item This survey collects accessible datasets, literature, and codes and makes them publicly available. Moreover, we analyze the potential applications of other deep learning techniques and structures in GVAED tasks.
  \item We examine the research challenges and trends within GVAED in the context of the development of deep learning techniques, which is anticipated to provide valuable guidance for upcoming researchers and engineers.
\end{enumerate}

The remainder of this survey is organized as follows. Section~\ref{sec2} provides an overview of the basics and research background of GVAED, including the definition of anomalies,  basic assumptions, main evaluation metrics, and benchmark datasets. Sections \ref{sec3}-\ref{sec6} introduce the unsupervised, weakly-supervised, supervised and fully-unsupervised GVAED methods, respectively. We analyze the extant methods' general ideas and specific implementations and compare their strengths and weaknesses. Further, we quantitatively compare the performance in  Section~\ref{sec7}. Section~\ref{sec8} analyzes the challenges and research outlook on the development of GVAED. Section~\ref{sec9} concludes this survey.


\section{Foundations of GVAED}~\label{sec2}
\subsection{Definition of the Anomaly}
UVAD follows the assumption of general AD tasks \cite{pang2021deep} and considers all events that have not occurred in the training set as abnormal. In other words, the training set of the UVAD dataset contains only normal events, while videos in the test set that differ from the training set are considered anomalies. Thus, certain normal events in the subjective human consciousness may also be labeled anomalies. For instance, in the UCSD Pedestrian datasets \cite{T2}, riding a bicycle on the college campus is labeled abnormal simply because the training set fails to contain such events. This seemingly odd definition is dictated by the diversity and rarity of real-world anomalies. Collecting a sufficient number of anomalous events with a full range of categories is nearly impossible. In response, researchers have taken the alternative route of collecting enough normal videos to train models to describe the boundary of normal patterns and treat events that fall outside the boundary as anomalies. Unfortunately, it is also costly to collect all possible normal events for training. In addition, abnormal and normal frames share most of the appearance and motion information, making their patterns overlap. Therefore, letting the model find a discriminative pattern boundary without seeing abnormal events is infeasible. 

In contrast, WAED takes a more intuitive definition of anomalies. Events that are subjectively perceived as abnormal by humans are considered anomalies, such as thefts and traffic accidents \cite{MIR}. The training set for WAED tasks contains both normal and abnormal events and provides easily accessible video-level labels to supervise the model. Compared with fine-grained frame-level labels, video-level labels only tell the model whether a video contains abnormal events without revealing the exact location of the abnormalities, avoiding the costly frame-by-frame labeling and providing more reliable supervision. In contrast, the discrete frame-level annotations (0=normal, 1=abnormal) in SVAD ignore the transition continuity from normal to abnormal events. WAED needs to predefine abnormal events so that it can only distinguish specified abnormal events. 

\begin{figure}
  \centering
  \includegraphics[width=\textwidth]{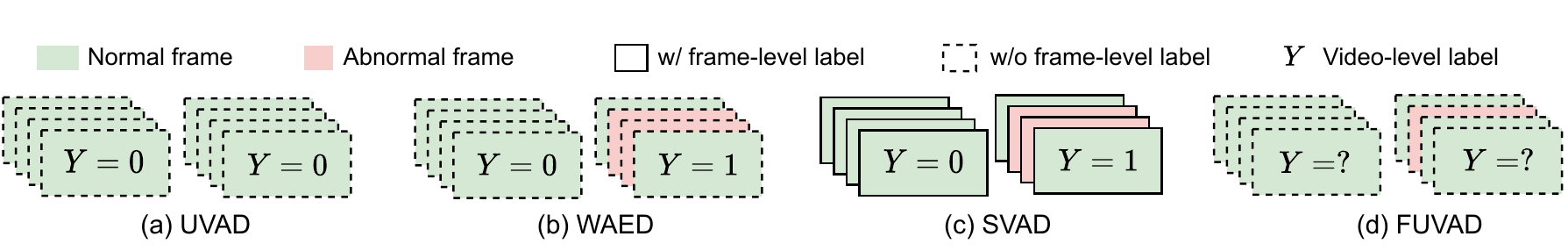}
  \caption{Illustration of training data. (a) UVAD trains the model using only normal data, with the hidden implication that all video-level and frame-level labels are 0. (b) WAED models use positive and negative samples and require frame-level labels, where $Y=0$ indicates normal video and $Y=1$ indicates an anomaly. (c) SVAD is trained using a fine-grained frame-level labeling supervised model, where the semantics of the frame-level labels expose the video-level labels. (d) FVAD attempts to learn the anomaly detector from under-processed data with training data containing both normal and anomalous samples and without any level of labeling. }
  \label{fig:xlsj}
\end{figure}

\subsection{Problem Formation}
\begin{figure}[t]
  \centering
  \includegraphics[width=\textwidth]{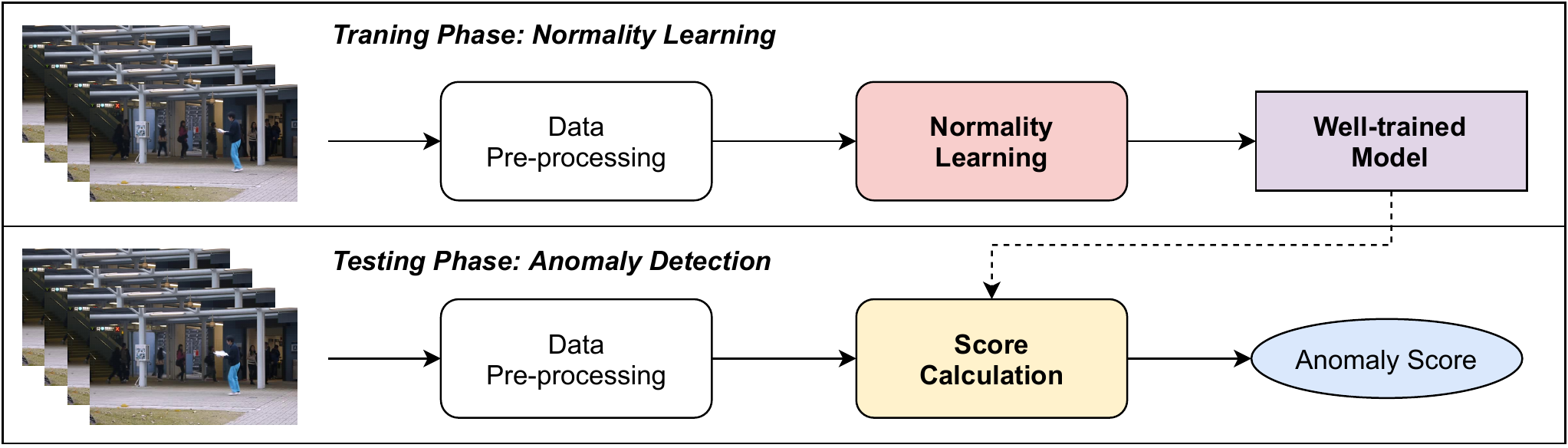}
  \caption{Illustration of the two-stage UVAD framework. Anomaly detection is performed in the test phase as a downstream task of proxy task-based normality learning. The example video frames are from the CUHK Avenue \cite{T3} dataset. }
  \label{fig:wjdkj}
\end{figure}

In UVAD, the training data contains only normal events, as shown in Fig.~\ref{fig:xlsj}(a). Such methods aim to describe the boundaries of normal spatial-temporal patterns with a proxy task and consider the test samples whose patterns fall outside the learned boundaries as anomalies \cite{liu2023stochastic}. 
Fig.~\ref{fig:wjdkj} shows the two-stage anomaly detection framework in UVAD. The deep network trained by performing the proxy task in the training phase is directly applied as a normality model for anomaly detection in the testing phase. The performance of the proxy task is a credential to calculate the anomaly score. Formulaically, the process of UVAD is as follows:
\begin{equation}
  e=d(f(\bm{x}_\text{test};\theta), \bm{x}_\text{test})
\end{equation}
where $\theta$ denotes the learnable parameters of the deep model $f$, designed to characterize the prototype of normal events. $d$ denotes the deviation between the test sample $x_\text{test}$ and the well-trained $f$, which is usually a quantifiable distance, such as the Mean Square Error (MSE) of the prediction result, the $L_2$ distance in the feature space and the difference of the distribution \cite{STAE,MNAD,FFPN}. Noting that the normality model is obtained by optimizing the proxy task. This process is independent of the downstream anomaly detection, so the performance of the proxy task cannot directly determine the anomaly detection performance. In addition, for the reconstruction-based \cite{FF-AE,memAE} and prediction-based \cite{FFP,STM-AE} methods, the final anomaly score is usually a relative value in the range $[0,1]$. A higher score indicates a larger deviation. Generally, these methods convert the absolute deviation $e$ into a relative anomaly score by performing maximum-minimum normalization. They not only explicitly require all training data to be normal but also include the hidden assumption that the test videos must include anomalous events. In other words, any test video will yield high score intervals, which indicates that such methods are offline and may produce false alarms for normal videos.

\begin{figure}
  \centering
  \includegraphics[width=\textwidth]{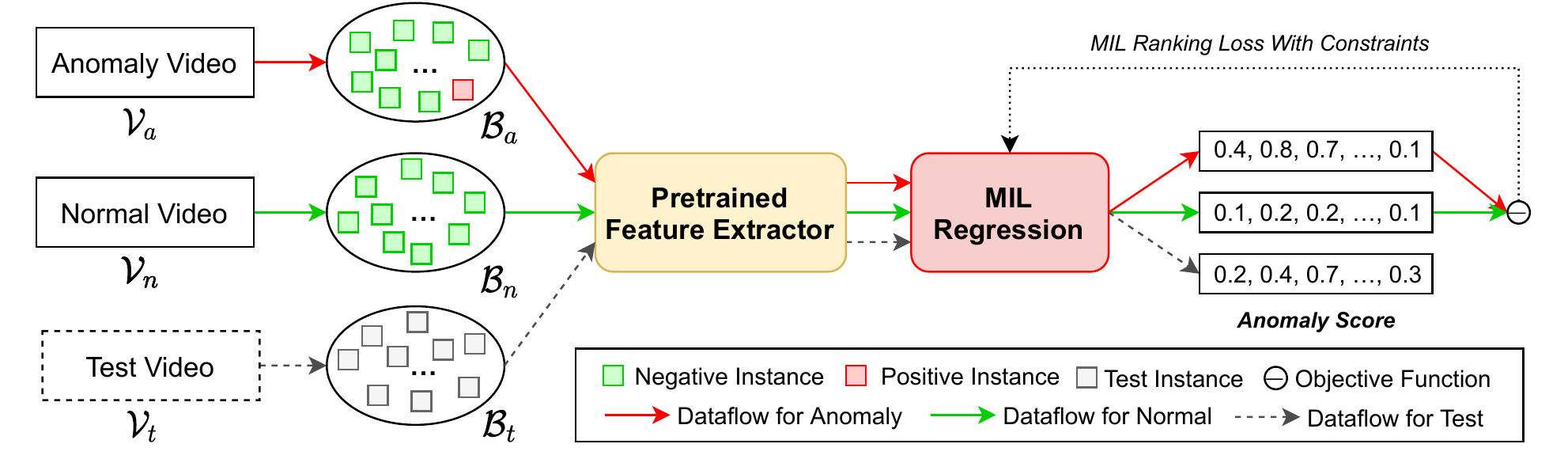}
  \caption{Structure of the MIL ranking model \cite{MIR}. The anomalous video $\mathcal{V}_a$ and the normal video $\mathcal{V}_n$ are first sliced into several equal-size instances. The positive bag $\mathcal{B}_a$ contains at least one positive instance, while the negative bag $\mathcal{B}_n$ contains only normal instances. In the test phase, the well-trained MIL regression model output the anomaly scores of instances in the test video $\mathcal{V}_t$ directly. }
  \label{fig:rjdkj}
\end{figure}

WAED methods \cite{MIST,RTFM,SMR} always follow the MIL ranking framework. Fig.~\ref{fig:xlsj}(b) shows the training data composition for WAED, where both normal and anomalous events need to be pre-collected and labeled. Video-level labels are easy to obtain and often more accurate than the fine-grained frame-level labels for SVAD shown in Fig.~\ref{fig:xlsj}(c). In a concrete implementation, WAED treats the video as a bag containing several instances, as illustrated in Fig.~\ref{fig:rjdkj}. The normal video $\mathcal{V}_n$ forms a negative bag $\mathcal{B}_n$, while the abnormal video $\mathcal{V}_a$ a positive bag $\mathcal{B}_a$. Based on MIL, WVEAD aims to train a regression model $r(\cdot)$ to assign scores to instances, with the basic goal that the maximum score of $\mathcal{B}_a$ is higher than that of $\mathcal{B}_n$. Thus, the WAED methods do not rely on an additional self-supervised proxy task but compute anomaly scores directly. The objective function $O\left(\mathcal{B}_a, \mathcal{B}_n\right)$ is as follows:
\begin{equation}
   O\left(\mathcal{B}_a, \mathcal{B}_n\right)=\min \max \left(0,1-\max _{i \in \mathcal{B}_a} r\left(\mathcal{V}_a^i\right)+\max _{i \in \mathcal{B}_n} r\left(\mathcal{V}_n^i\right)\right) 
  +\lambda_1 \overbrace{\sum_i^{n-1}\left(r\left(\mathcal{V}_a^i\right)-r\left(\mathcal{V}_a^{i+1}\right)\right)^2}^{C_{smooth}}+\lambda_2 \overbrace{\sum_i^n r\left(\mathcal{V}_a^i\right)}^{C_{sparsity}}
\end{equation}
In addition to the additional anomaly curve smoothness constraint $C_{smooth}$ and sparsity constraint $C_{sparsity}$, the core of $O\left(\mathcal{B}_a, \mathcal{B}_n\right)$ is to train a ranking model capable of distinguishing the spatial-temporal patterns between $\mathcal{B}_a$ and $\mathcal{B}_n$. Subsequent WAED works \cite{MAF,ARNet,RTFM,MIST,SMR,STA} have followed the idea of MIL ranking and made effective improvements regarding feature extraction \cite{MAF}, label denoising \cite{GCLNC}, and the objective function \cite{SMR}. However, as shown in Fig.~\ref{fig:rjdkj}, the MIL regression module takes the extracted feature representations as input, so the performance of WAED methods partially depends on the pre-trained feature extractor, making the calculation costly. 

The training data of FVAD contains both normal and abnormal, and none of the data labels are available for model training, as shown in Fig.~\ref{fig:xlsj}(d). One class of FVAD methods follows a similar workflow to that of UVAD, i.e., learning the normality model directly from the original data. Although the training data contains anomalous events, the low frequency of anomalies limits their impact on model optimization. As a result, the model learned on many normal videos and a small number of abnormal frames is still only effective in representing normal events and generates large errors for anomalous events. Another class of methods tries to discover anomalies through the mutual collaboration of the representation learner and anomaly detector. Generally, the learning process of FVAD can be formulated as follows:
\begin{equation}
  \mathcal{F}=\underset{\Theta}{\arg \min } \sum_{I \in \mathbb{I}} \mathcal{L}_{f o c}(\hat{y}=\phi(m=\varphi(x=f(I))), l)
  \end{equation}
  where the aim is to learn an anomaly detector $\mathcal{F}$ via a deep neural network which consists of a backbone network $f\left(\cdot ; \Theta_b\right): \mathbb{R}^{H \times W \times 3} \mapsto \mathbb{R}^{D_b}$ that transforms an input video frame $I$ to feature $\boldsymbol{x}$, an anomaly representation learner $\varphi\left(\cdot ; \Theta_a\right): \mathbb{R}^{D_b} \mapsto \mathbb{R}^{D_n}$ that converts $x$ to an anomaly specific representation $m$, and an anomaly score regression layer $\phi\left(\cdot ; \Theta_s\right): \mathbb{R}^{D_s} \mapsto \mathbb{R}$ that learns to predict $m$ to an anomaly score $y$. The overall parameters $\Theta=\left\{\Theta_b, \Theta_a, \Theta_s\right\}$ are optimized by the focal loss.
Research on fully-unsupervised methods is still in its infancy, and they exploit the imbalance of samples and the significant difference of anomalies in the GVAED task.

\subsection{Benchmark Datasets}
In Table~\ref{tab:sjj}, we show and compare the statistical results and properties of the frequently used GVAED datasets. Several datasets \cite{MIR,HL-Net,FFP,UBnormal} have been proposed with different annotated signals to match new research requirements after 2018, which reflects the trend of GVAED from unsupervised to weakly-unsupervised \cite{MIR}, from unimodal to multimodal \cite{HL-Net} and from simple to complex real-world scenarios \cite{FFP} at the data level. 

\subsubsection{Subway Entrance \& Exit}
As an earlier dataset, Subway \cite{Subway} includes two independent sub-datasets, Entrance and Exit, which record the subway entrance and exit scenes, respectively. 
The anomalous events include people who skip the subway entrance to evade tickets, cleaners who behave differently from regular entry and exit, and people who travel in the wrong direction. Due to the cursory nature of the labeling work and the lack of clarity in the definition of anomalous events, most existing works have refrained from using this dataset for model evaluation. Therefore, we do not provide quantitative performance comparison results on this dataset but only briefly describe its characteristics to reflect the lineage of GVAED datasets development.

\begin{table}[]
  \centering
  \caption{Representative GVAED Datasets. \textit{Italicized ones }indicate WAED datasets, and \underline{underlined one} is multimodal dataset.}
  \label{tab:sjj}
  \begin{threeparttable}
  \resizebox{\textwidth}{!}{
  \begin{tabular}{@{}clccc|ccccccc@{}}
  \toprule
  \multirow{2}{*}{\textbf{Year}} & \multicolumn{1}{c}{\multirow{2}{*}{\textbf{Dataset}}} & \multicolumn{3}{c|}{\textbf{\#Videos}} & \multicolumn{5}{c}{\textbf{\#Frames}}  & \multirow{2}{*}{\textbf{\#Scenes}} & \multirow{2}{*}{\textbf{\#Anomalies}} \\ \cmidrule(lr){3-10}
     & \multicolumn{1}{c}{}      & Total      & Training     & Testing & Total      & Training   & Testing   & Normal  & Abnormal &  &  \\ \midrule
  2008     & Subway Entrance\tablefootnote{\url{https://vision.eecs.yorku.ca/research/anomalous-behaviour-data/sets/}}  & -  &  - &  - & 144,250    & 76,543     & 67,797    & 132,138 & 12,112   & 1      & 51  \\
  2008     & Subway Exit\footnotemark[2]     &  - &  - & -  & 64,901     & 22,500     & 42,401    & 60,410  & 4,491    & 1      & 14  \\
  2011     & UMN\tablefootnote{\url{http://mha.cs.umn.edu/proj_events.shtml\#crowd}}$^\dagger$  & -  &  -   &  - & 7,741      &   -   &  -   & 6,165   & 1,576    & 3      & 11  \\
  2013     & UCSD Ped1\tablefootnote{\url{http://www.svcl.ucsd.edu/projects/anomaly/dataset.htm}}  & 70   & 34  & 36      & 14,000     & 6,800      & 7,200     & 9,995   & 4,005    & 1      & 61  \\
  2013     & UCSD Ped2\footnotemark[4]  & 28   & 16  & 12      & 4,560      & 2,550      & 2,010     & 2,924   & 1,636    & 1      & 21  \\
  2013     & CUHK Avenue\tablefootnote{\url{http://www.cse.cuhk.edu.hk/leojia/projects/detectabnormal/dataset.html}}
  & 37   & 16  & 21      & 30,652     & 15,328     & 15,324    & 26,832  & 3,820    & 1      & 77  \\
  2018     & ShanghaiTech\tablefootnote{\url{https://svip-lab.github.io/dataset/campus_dataset.html}}  &  - &  -   &  - & 317,398    & 274,515    & 42,883    & 300,308 & 17,090   & 13     & 158       \\
  2018     & \textit{UCF-Crime}\tablefootnote{\url{https://webpages.charlotte.edu/cchen62/dataset.html}}  & 1,900      & 1,610  & 290     & 13,741,393 & 12,631,211 & 1,110,182 &  - & -   & - & 950       \\
  2019     & \textit{ShanghaiTech Weakly}\tablefootnote{\url{https://github.com/jx-zhong-for-academic-purpose/GCN-Anomaly-Detection/}}       & 437  & 330 & 107     &  -    &  -    &  -   &  - &  -  &-  &-  \\
  2020     & Street Scene\tablefootnote{\url{https://www.merl.com/demos/video-anomaly-detection}}  & 81   & 46  & 35      & 203,257    & 56,847     & 146,410   & 159,341 & 43,916   & 205    & 17  \\
  2020     & \underline{\textit{XD-Violance}}\tablefootnote{\url{https://roc-ng.github.io/XD-Violence/}}  & 4,754      &   -  &  - &  -    &  -    & -    & -  &  -  & - & - \\
  2022     & UBnormal\tablefootnote{\url{https://github.com/lilygeorgescu/UBnormal}} $^\ddagger$  & 543  & 268 & 211     & 236,902    & 116,087    & 92,640    & 147,887 & 89,015   & 29     & 660       \\ \bottomrule
  \end{tabular}}
  \begin{tablenotes}    
    \footnotesize   
    \item $^\dagger$ Following previous works, we set the frame rate to 15 fps. $\ddagger$ The UBnormal dataset is supervised and includes a validation set with 64 videos.
  \end{tablenotes}   
\end{threeparttable}
  \end{table}

\subsubsection{UMN}
The UMN \cite{UMN} is also an early GVAED dataset containing 11 short videos captured from three different scenes: grassland, indoor hall, and park. 
The scenes are set by the researcher rather than naturally filmed to detect abnormal crowd behavior in indoor and outdoor scenes, i.e., the crowd suddenly shifts from normal interaction to evacuation and flees abruptly to simulate fear. The anomalies are artificially conceived and played out, ignoring the diversity and rarity of anomalies in the real-world. Similar to the Subway \cite{Subway} dataset, UMN has been abandoned by recent researchers due to the lack of spatial annotation.

\subsubsection{UCSD Pedestrian}
UCSD Ped1 \& Ped2 \cite{T2} are the most widely used UVAD datasets collected from university campuses with simple but realistic scenarios. They reflect the value of GVAED in public security applications. Specifically, the Ped1 dataset is captured by a camera with a viewpoint perpendicular to the road, so the moving object's size changes with its spatial position. In contrast, the Ped2 dataset used a camera whose viewpoint is parallel to the direction of the road, which is simpler than Ped1. Pedestrian walking is defined as normal, while behaviors and objects different from it are considered abnormal, such as biking, skateboarding, and driving. Since the scene is classical and anomalous events are easy to understand, UCSD Pedestrian is widely used by existing works, and the frame-level AUC has been as high as 99\%, reflecting the saturation of model performance. Therefore, the dataset in simple scenes has become a constraint for GVAED development. The large-scale and cross-scene datasets have become an inevitable trend.

\subsubsection{CUHK Avenue}

Similar to UCSD Pedestrian, the CUHK Avenue \cite{T3} dataset is also collected from the university campus, and both focus on anomalous events that occur on the road outside of expectations. The difference is that most of the 47 anomalous events in CUHK Avenue are simulated by the data collector, including appearance anomalies (e.g., bags placed on the grass) and motion anomalies, such as throwing and wrong direction. CUHK Avenue provides both frame-level and pixel-level spatial annotations. In addition, its large data scale makes it one of the mainstream UVAD datasets.

\subsubsection{ShanghaiTech}

The UCSD Pedestrian \cite{T2} and CUHK Avenue \cite{T3} datasets only consider anomalous events in a single scene, while the real world usually faces the challenge of spatial-temporal pattern shifts across scenes. For this reason, the team from ShanghaiTech University proposed the ShanghaiTech \cite{FFP} dataset containing 13 scenes, providing the largest UVAD  benchmark. Abnormal behaviors are defined as all collected behaviors that distinguish them from normal walking, such as riding a bicycle, crossing a road, and jumping forward.
Unfortunately, although the collectors pointed out the shortcomings of the existing dataset with a single scenario, their proposed FFP \cite{FFP} was not explicitly designed to address the cross-scene challenges but rather to treat it as a whole without differentiating between scenarios. For the WAED setting, researchers \cite{GCLNC} proposed to move some anomalous videos from the test set to the training set and provided video-level labels for each training video, introducing the ShanghaiTech Weakly dataset, which has become the mainstream WAED benchmark. A compelling phenomenon is that the performance of WAED methods on ShanghaiTech Weakly (frame-level AUC is typically $>$ 85\% and has reached up to 95\%) is generally higher than that of UVAD methods on the ShanghaiTech (frame-level AUC is typically between $70\sim 80$\%), providing evidence for the applicability of WAED in complex scenarios over UVAD.

\subsubsection{UCF-Crime}

UCF-Crime \cite{MIR} is the first WAED dataset, presented together with the original MIL ranking framework. UCF-Crime consists of 1900 unedited real-world surveillance videos collected from the Internet. The abnormal events contain 850 anomalies of human concern in 13 categories: \textit{Abuse}, \textit{Arrest}, \textit{Arson}, \textit{Assault}, \textit{Burglary}, \textit{Explosion}, \textit{Fighting}, \textit{Road Accidents}, \textit{Robbery}, \textit{Shooting}, \textit{Shoplifting}, \textit{Stealing}, and \textit{Vandalism}. Unlike the UVAD dataset above, its training set contains anomalous videos and provides a video-level label for each video, where 0 indicates normal, and 1 indicates anomalous. The anomalous events in the WAED dataset are predefined and are usually associated with specific scenarios, such as car accidents in urban traffic, shoplifting, and shootings in neighborhoods.
Therefore, WAED can provide more credible results for real scenarios with better application potential.

\subsubsection{XD-Violence}
\begin{figure}
  \centering
  \includegraphics[width=\textwidth]{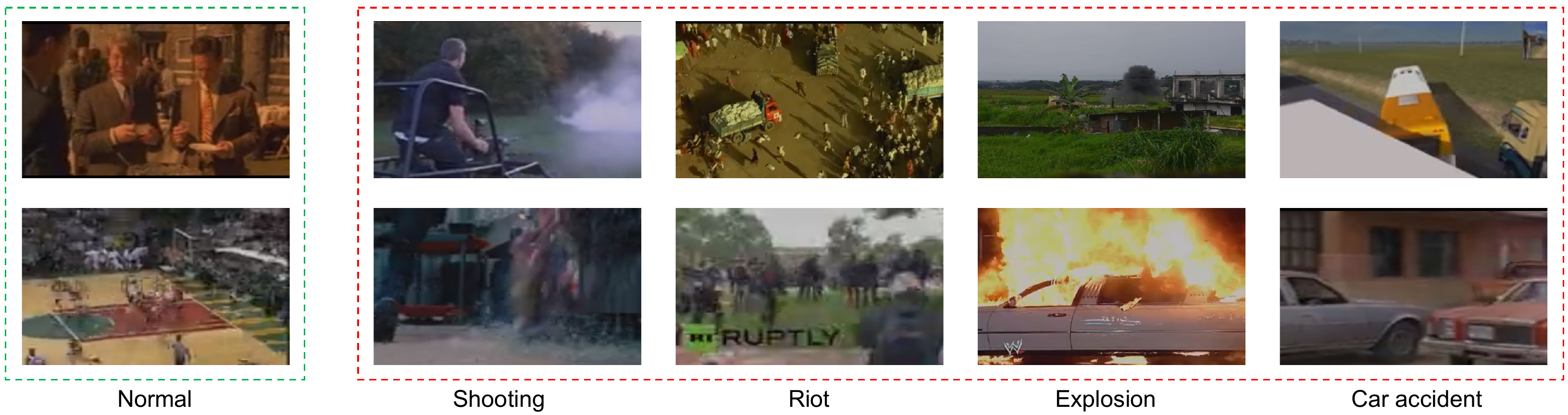}
  \caption{Examples of XD-Violence dataset \cite{HL-Net}. XD-Violence is a multimodal dataset for violence detection, including video and audio. We show video frames here. The anomalous events are not all from the real-world but also include movie and game footage, etc.}
  \label{fig:xd}
\end{figure}
As the first audio-video dataset, XD-Violence \cite{HL-Net} expands anomaly event detection from single-modal video understanding to multimodal signal processing, facilitating the coexistence of GVAED and multimedia communities. XD-Violence focuses on violent behaviors, such as abuse, explosion, car accident, struggle, shootings, and riots, as shown in Fig.~\ref{fig:xd}. Due to the rarity of violent behaviors and the high difficulty of capturing violence, the original videos include some movie clips in addition to real-world surveillance videos. XD-Violence provides a new way to think about the GVAED by extending the data modality from single videos to sound, text, and others.

\subsubsection{UBnormal}

Inspired by the computer vision community benefiting from synthetic data, Acsintoae \textit{ et al.} \cite{UBnormal} propose the first GVAED benchmark with virtual scenes, named UBnormal. Notably, utilizing a data engine to synthesize data under predetermined instructions rather than collecting real-world data makes pixel-level labeling possible. Therefore, UBnormal is supervised. UBnormal is built to address the problem that WAED ignores the open-set nature of anomalies that prevents the model from correctly corresponding to new anomalies. The test set contains anomalous events not present in the training set. Moreover, it provides a validation set for model tuning for the first time.

\begin{figure}
  \centering
  \includegraphics[width=.95\textwidth]{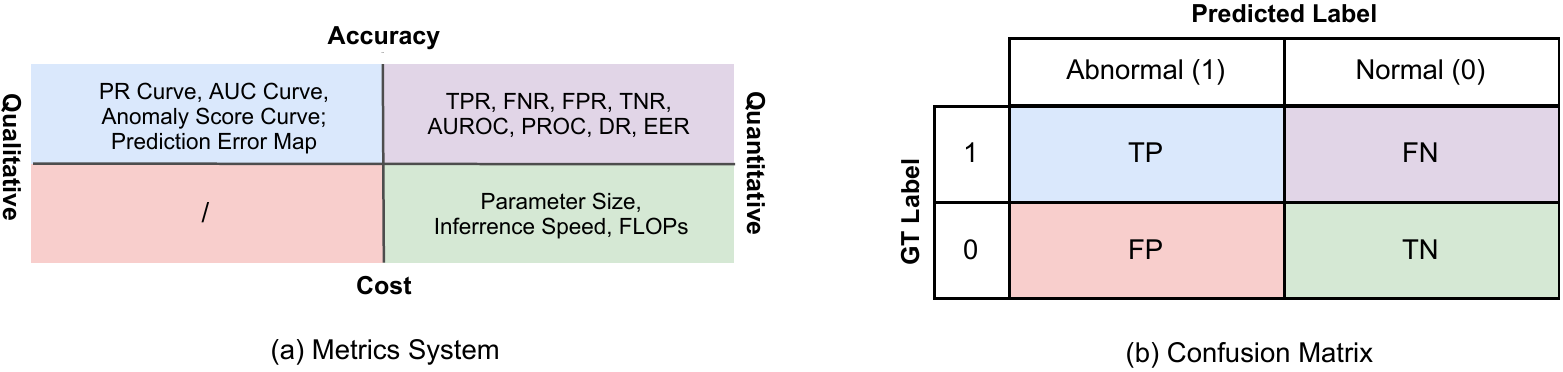}
  \caption{Illustration of GVAED performance evaluation system. We show the (a) metrics system and (b) confusion matrix.}
  \label{fig:xnpg} 
\end{figure}

\subsection{Performance Evaluation}
Existing GVAED methods evaluate model performance in terms of detection accuracy and operational cost. The former concerns the ability to discriminate anomalous events while the latter aims to measure the deployment potential on resource-limited devices. According to the scale of detected anomalies, the detection accuracy criteria are divided into three levels: Temporal-Detection-Oriented (TDO), Object-Detection-Oriented (ODO), and Spatial-Localization-Oriented (SLO). Specifically, TDO criteria require the model to determine anomalous events' starting and ending temporal position without spatial localization of abnormal pixels. In contrast, ODO criteria include object-level, region-level, and track-level, focusing on specific anomaly objects or trajectories. SLO criteria encourage pixel-level localization of anomalous events. As for operational cost criteria, the commonly used metric include parameter size, inference speed, and the number of FLOating Point operations (FLOPs) on the same platform, as shown in Fig.~\ref{fig:xnpg}(a).

We can evaluate the model performance quantitatively by comparing the predicted results with the ground truth labels. It is worth noting that the predicted labels of some GVAED models (e.g., prediction-based UVAD and WAED) are continuous values in the range of $[0, 1]$. In contrast, the true labels are discrete 0 or 1, so a threshold value must first be selected when calculating the performance metrics. Samples with abnormal scores below the threshold are considered normal, and vice versa. In this way, we obtain the confusion matrix shown in Fig.~\ref{fig:xnpg}(b), where TP, FN, FP, and TN denote the number of abnormal samples correctly detected, abnormal samples mistakenly detected as normal, normal samples mistakenly detected as abnormal, and normal samples correctly detected, respectively. The True-Positive-Rate (TPR), False-Positive-Rate (FPR), True-Negative-Rate (TNR), and False-Negative-Rate (FNR) are defined as follows:
\begin{equation}
  \label{eq:tpr}
  \text{TPR}=\frac{\text{TP}}{\text{TP+FN}};  \text{FPR}=\frac{\text{FP}}{\text{FP+TN}}; \text{TNR}=\frac{\text{TN}}{\text{FP+TN}}; \text{FNR}=\frac{\text{FN}}{\text{TP+FN}}
\end{equation}

which are used to calculate the Area Under the Receiver Operating Characteristic (AUROC) and Average Precision (AP).

\textbf{AUROC: }The horizontal and vertical coordinates of the Receiver Operating Characteristic (ROC) curve are the FPR and TPR, and the curve is obtained by calculating the FPR and TPR under multiple sets of thresholds. The area of the region enclosed by the ROC curve and the horizontal axis is often used to evaluate binary classification tasks, denoted as AUROC. The value of AUROC is within the range of [0, 1], and higher values indicate better performance. AUROC can visualize the generalization performance of the GVAED model and help to select the best alarm threshold. In addition, the Equal Error Rate (EER), i.e., the proportion of incorrectly classified frames when TPR and FNR are equal, is also used to measure the performance of anomaly detection models.

\textbf{AP:} Due to the highly unbalanced nature of positive and negative samples in GVAED tasks, i.e., the TN is usually larger than the TP, researchers think that the area under the Precision-Recall (PR) curve is more suitable for evaluating GVAED models, denoted as AP. The horizontal coordinates of the PR curve are the Recall (i.e., the TPR in Eq.~\ref{eq:tpr}), while the vertical coordinate represents the Precision, defined as $\text{Precision}=\frac{\text{TP}}{\text{TP+FP}}$. A point on the PR curve corresponds to the Precision and Recall values at a certain threshold. Currently, AP has become the main metric for multimodal GVAED models \cite{HL-Net,HL-Net+,MSAF} and is widely used to evaluate the performance on the XD-Violence dataset \cite{HL-Net}.

\section{Unsupervised Video Anomaly Detection}~\label{sec3}

Existing reviews \cite{nayak2021comprehensive,ramachandra2020survey} usually classify UVAD methods into distance-based \cite{T68,LGA,T69}, probability-based \cite{T49,T4,T40,T52}, and reconstruction-based \cite{STM-AE,FF-AE,memAE} according to the deviations calculation means. 
Early traditional methods relied on manual features such as foreground masks \cite{LGA}, histogram of flow \cite{T68}, motion magnitude \cite{LGA}, HOG \cite{T69}, dense trajectories \cite{T71}, and STIP \cite{T72}, which relied on human a priori knowledge and had poor representational power. With the rise of deep learning in computer vision tasks \cite{r1,r2,r3}, recent approaches preferred to extracting features representations in an end-to-end framework with deep Auto-Encoder (AE) \cite{FF-AE,T18,AMAE,STM-AE,MNAD}, Generative Adversarial Network (GAN) \cite{chen2021nm,TAC-Net,nguyen2020anomaly,li2022context,cai2021video,zhang2022surveillance,huang2022self}, and Vision Transformer (ViT) \cite{CT-D2GAN,yuan2021transanomaly,lee2022multi}. 

This section is dedicated to providing a systematic overview of UVAD methods driven by deep learning techniques \cite{deng2023bi,liu2023msn,cheng2023spatial,liu2023learning,cheng2023learning}. It's worth noting that the traditional taxonomy, as outlined in previous studies \cite{nayak2021comprehensive,ramachandra2020survey}, predominantly focuses on manual feature-based methods, which are limited in elucidating the evolving landscape of deep UVAD models. Deep CNNs exhibit a remarkable capacity for modeling the spatiotemporal intricacies within video sequences and generating profound representations of various sensory domains, contingent upon the nature of the input data. Consequently, we categorize the work found in existing literature into three principal groups, contingent upon the nature of the data employed: 1) \textbf{Frame-level methods} always utilize entire RGB and optical flow frames as input and endeavor to capture a holistic understanding of normality within the video data. Such approaches consider the entirety of the frame, aiming to comprehend the global context. 2) \textbf{Patch-level methods} recognize the repetitive spatial-temporal information present in video sequences, so they extract features solely from designated regions of interest. They intentionally disregard redundant data from repetitive regions and interactions among regional information that do not warrant particular attention. This strategy offers distinct advantages in terms of computational efficiency and inference speed. 3) \textbf{Object-level methods}, emerging in recent years with the development of target detection models, shift the focus towards detecting foreground objects and scrutinizing the behavior of specific objects within the video context. Object-level methods consider the relationship between objects and their backgrounds, showcasing impressive performance in the task of identifying anomalous events within complex scenes. Based on the aforementioned analysis, this section classifies UVAD methods into three distinct categories: frame-level, patch-level, and object-level. This categorization is aligned with a hierarchical \textit{"input-structure"} taxonomy shown in Fig.~\ref{fig:overview}. This taxonomy serves as a guiding framework for organizing and understanding the landscape of UVAD.

\subsection{Frame-level Methods}
Deep CNNs can directly extract abstract features from videos and learn task-specific deep representations. Frame-level methods use complete RGB frames, sequences, or optical flows as input to model the normality of normal events in a self-supervised learning manner. Existing methods can be classified into two categories according to model structure: single-stream and multi-stream. The former does not distinguish spatial and temporal information. They usually take the original RGB videos as input and learn the spatial-temporal patterns by reconstructing the input sequence or predicting the next frame. Existing single-stream work focuses on designing more efficient network structures. They introduce more powerful representational learners such as 3D convolution \cite{STAE} and U-net \cite{FFP}. In contrast, multi-stream networks typically treat appearance and motion as different dimensions of information and attempt to learn spatial and temporal normality using different agent tasks or network architectures. In addition to spatial-temporal separation modeling, existing dual-stream works explored spatial-temporal coherence \cite{AMAE} and consistency \cite{AMMC-net} to perform anomaly detection. 

\subsubsection{Single-Stream Models} Single-stream models typically use a single generative model to describe the spatial-temporal patterns of normal events by performing a proxy task and preserving the normality in learnable parameters. For example, the Predictive Convolutional Long Short-Term Memory (PC-LSTM) \cite{PC-LSTM} used a conforming ConvLSTM network to model the evolution of video sequences. Hasan \textit{et al.} \cite{FF-AE} constructed a fully convolutional Feed-Forward Auto-Encoder (FF-AE) with manual features as input, which can learn task-specific representations in an end-to-end manner. 

Liu \textit{et al.} \cite{FFP} proposed a Future Frame Prediction (FFP) method that used a GAN-based video prediction framework to learn the normality. Its extension, FFPN \cite{FFPN}, further specified the design principles of predictive UVAD networks. Singh and Pankajakshan \cite{LSTM-AE} also used a predictive task to detect anomalies, proposing conformal structures based on 2D \& 3D convolution and convLSTM to characterize spatial-temporal patterns more efficiently. 

To address the detail loss in frame generation, Li \textit{et al.} \cite{STU-net} proposed a Spatial-Temporal U-net network (STU-net) that combined the advantages of U-net in representing spatial information with the ability of convLSTM to model temporal variations for moving objects. \cite{AnomalyNet} proposed a sparse coding-based neural network called AnomalyNet, which used three neural networks to integrate the advantages of feature learning, sparse representation, and dictionary learning. In \cite{ISTL}, the authors proposed an Incremental Spatial-Temporal Learner (ISTL) to explore the nature of anomalous behavior over time. ISTL used active learning with fuzzy aggregation to continuously update and distinguish between new anomalous and normal events evolving. The anoPCN in \cite{AnoPCN} unified the reconstruction and prediction methods into a deep predictive coding network by introducing an error refinement module to reconstruct the prediction errors and refining the coarse predictions generated by the predictive coding module.

To lessen the deep model's ability to generalize anomalous samples, memory-augmented Auto-Encoder (memAE) \cite{memAE} embedded an external memory network between the encoder and decoder to record the prototypical patterns of normal events. Further, Park \textit{et al.} \cite{MNAD} introduced an attention-based memory addressing mechanism and proposed to update the memory pool during the testing phase to ensure that the network can better represent normal events.

Luo \textit{et al.} \cite{TSC} proposed a sparse coding-inspired neural network model, namely Temporally-coherent Sparse Coding (TSC). It used a sequential iterative soft thresholding algorithm to optimize the sparse coefficients. \cite{R-STAE} introduces residual connection \cite{Resnet} into the auto-encoder to eliminate the gradient disappearance problem during normality learning. Experiments have shown that ResNet brong 3\%, 2\% and 5\% frame-level AUC gains for the proposed Residual Spatial-Temporal Auto-Encoder (R-STAE) on CUHK Avenue \cite{T3}, LV \cite{LV} and UCSD Ped2 \cite{T2} datasets, respectively.

The DD-GAN in \cite{DD-GAN} introduced an additional motion discriminator to GAN. The dual discriminators structure encouraged the generator to generate more realistic frames with motion continuity. Yu \textit{et al.} \cite{AEP} also used GAN to model normality. The proposed Adversarial Event Prediction (AEP) network performed adversarial learning on past and future events to explore the correlation. Similarly, Zhao \textit{et al.} \cite{STC-Net} explored spatial-temporal correlations by GAN and used a spatial-temporal LSTM to extract appearance and motion information within a unified unit.

\cite{Bi-Pre} proposed a Bidirectional Prediction (Bi-Pre) framework that used forward and backward prediction sub-networks to reason about normal frames. In the test phase, only part significant regions are used to calculate the anomaly score, allowing the model to focus on the foreground. Wang \textit{et al.} \cite{ROADMAP} used multi-path convGRU to perform frame prediction. The proposed ROADMAP model included three non-local modules to handle different scales of objects. 

\begin{table}[t]
  \centering
  \caption{Frame-level Multi-stream UVAD Methods.}
  \label{tab:slff}
  \resizebox{\textwidth}{!}{
 \begin{tabular}{@{}cp{2cm}p{2.2cm}p{9.1cm}@{}}
\toprule
\multicolumn{1}{c}{\textbf{Year}} & \multicolumn{1}{c}{\textbf{Method}} & \multicolumn{1}{c}{\textbf{Backbone}} & \multicolumn{1}{c}{\textbf{Analysis}}  \\ \midrule
2017 & AMDN \cite{AMDN}     & AE       & \makecell[l]{\textbf{Pros:} Learning appearance and motion patterns separately.\\\textbf{Cons:} Determining boundaries by OC-SVM with limited capability.}    \\ \midrule
2017 & STAE \cite{STAE}    & AE       & \makecell[l]{\textbf{Pros: }Using 3D CNN to learn the spatial-temporal patterns.\\\textbf{Cons: }Dual decoders causing huge computational costs.}   \\ \midrule
2019 & AMC \cite{AMC}     & GAN      & \makecell[l]{\textbf{Pros: }Learning the correspondence between appearance and motion.\\\textbf{Cons: }Limited performance on the complex datasets.}         \\ \midrule
2019 & GANs \cite{GANs}    & GAN      & \makecell[l]{\textbf{Pros:} Training two GANs to learn temporal and spatial distribution.\\\textbf{Cons: }Unstable traning process and high training cost.}       \\ \midrule
2020 & CDD-AE \cite{CDD-AE}  & AE       & \makecell[l]{\textbf{Pros: }Using two auto-encoders to learn spatial and temporal patterns.\\\textbf{Cons: }No special consideration in the design of the encoder structure.}    \\ \midrule
2020 & OGNet \cite{OGNet}   & GAN      & \makecell[l]{\textbf{Pros: }Using generators and discriminators to learn normality.\\\textbf{Cons: }Adversarial learning making the training process unstable.}        \\ \midrule
2021 & AMMC-net \cite{AMMC-net} & AE       & \makecell[l]{\textbf{Pros:} Exploring the consistency of appearance and motion.\\\textbf{Cons: }  Lack of analysis to the flow-frame generation task.}  \\ \midrule
2021 & DSTAE \cite{DSTAE}    & AE, ConvLSTM   & \makecell[l]{\textbf{Pros: }Using two auto-encoders to perform different tasks. \\\textbf{Cons: }High computional cost and relying on optical flow network.}  \\ \midrule
2022 & AMAE \cite{AMAE}    & AE       & \makecell[l]{\textbf{Pros: }Using two encoders and three decoders to learn features. \\\textbf{Cons: } High training cost and relying on optical flow network.}  \\ \midrule
2022 & STM-AE \cite{STM-AE}  & AE, GAN  & \makecell[l]{\textbf{Pros: }Using two memory-enhanced auto-encoders to learn normality. \\\textbf{Cons: }Unstable traning process and high computional costs.}      \\ \bottomrule
\end{tabular}}
\end{table}
\subsubsection{Multi-Stream Models}. The \textit{multiplicity} of multi-stream models is reflected in the multiple sources of the input data and the multiple tasks corresponding to multiple outputs. Considering that video anomaly may manifest as outliers in appearance or motion, an intuitive idea is to use multi-stream networks to model spatial and temporal normality separately \cite{CDD-AE,CDD-AE+,DSTAE,wang2023memory,wu2023dss}. In addition, learning associations between appearance and motion, such as consistency\cite{AMMC-net}, coherence \cite{AMAE,STM-AE}, and correspondence \cite{AMC}, is another effective GVAED solution. Events without such associations are discriminated against as anomalies. The multi-stream model has achieved significant success in recent years due to the high matching of its design motivation with the GVAED task. The multi-stream methods are summarized in Table~\ref{tab:slff}.

Motivated by the remarkable success of 3D CNN in video understanding tasks, Zhao \textit{et al.} \cite{STAE} proposed a 3D convolutional-based Spatial-Temporal Auto-Encoder (STAE) to model normality by simultaneously performing reconstruction and prediction tasks. STAE included two decoders, which outputted reconstructed and predicted frames, respectively. In contrast, Appearance and Motion DeepNet (AMDN) \cite{AMDN} used two stacked denoising auto-encoders to encode RGB frames and optical flow separately. Similarly, Chang \textit{et al.} \cite {CDD-AE} also used two auto-encoders to capture spatial and temporal information, respectively. One learned the appearance by reconstructing the last frame, while the other outputted RGB differences to simulate the generation of optical flow. Deep K-means clustering was used to force the extracted feature compact and detect anomalies. DSTAE \cite{DSTAE} introduced convLSTM to a two-stream auto-encoder to better model the temporal variations. The reconstruction errors of the two encoders are weighted and used to calculate anomaly scores.

In addition to spatial-temporal separation, Nguyen and Meunier \cite{AMC} proposed to learn the correspondence between appearance and motion. To this end, they proposed an AE with two decoders, one for reconstructing input frames and the other for predicting optical flow. Cai \textit{et al.} \cite{AMMC-net} proposed an Appearance-Motion Memory Consistency network (AMMC-net), which aimed to capture the spatial-temporal consistency in high-level feature space.

Liu \textit{et al.} \cite{AMAE} proposed an Appearance-Motion united Auto-Encoder (AMAE) framework using two independent auto-encoders to perform denoising and optical flow generation tasks separately. Moreover, they utilized an additional decoder to fuse spatial-temporal features and predict future frames to model spatial-temporal normality. STM-AE \cite{STM-AE} and AMP-Net \cite{amp} introduced the memory into the dual-stream auto-encoder to record prototype appearance and motion patterns. Adversarial learning was used to explore the connection between spatial and temporal information of regular events.

Aside from the above anomaly detection means such as reconstruction error \cite{STAE,DSTAE,AMAE,STM-AE}, clustering \cite{CDD-AE, CDD-AE+} and one-class classification \cite{AMDN}, researchers attempted to utilize the discriminator of GAN to directly output results. For instance, Ravanbakhsh \textit{et al.} \cite{GANs} used GAN to learn the normal distribution and detect anomalies directly by discriminators. The authors use a cross-channel approach to prevent the discriminator from learning mundane constant functions. OGNet \cite{OGNet} shifted the discriminator from discriminating real or generated frames to distinguishing good or poor reconstructions. The well-trained discriminator can find subtle distortions in the reconstruction results and detect non-obvious anomalies.

The patch-level methods \cite{ADCS,STC,Deep-Cascade,AST-AE} takes the video patch (spatial-temporal cube) as input. Compared with frame-level methods that consider anomalies roughly, i.e., anomalies are reflected in spatial or temporal dimensions beyond expectation, patch-level methods consider finding anomalies from specific spatial-temporal regions rather than analyzing the whole sequence. Patch formation can be divided into three categories: scale equipartition \cite{ADCS,STCNN,DeepAnomaly,Deep-Cascade,DeepOC,ST-CaAE,AST-AE}, information equipartition \cite{T108}, and foreground object extraction \cite{S2-VAE}. Specifically, scale equipartition is the simplest. The video sequence is equipartitioned into several spatial-temporal cubes of uniform size along the spatial and temporal dimensions. The subsequent modeling process is similar to frame-level methods. The information equipartition strategy considers that image blocks of the same size do not contain the same information. Regions close to the camera contain less information per unit area than those far away. Before representation, all cubes will be first resized to the same size. The foreground object extraction focuses on modeling regions with information variation to avoid the learning cost and disruption of the background. After the sequences are equated into same-scale cubes, those containing only background will be eliminated. 

Roshtkhari and Levine \cite{STC} densely sampled video sequences at different spatial and temporal scales and used a probabilistic framework to model the spatial-temporal composition of the video volumes. The STCNN \cite{STCNN} treated UVAD as a binary classification task. It first extracted patches' appearance and motion information and outputted the discriminative results with an FCN. It first equated the video sequence into patches of $3\times 3\times 7$ and retained only the part of the region containing moving pixels to ensure the robustness of the model to local noise and improve the detection accuracy. Deep-Cascade \cite{Deep-Cascade} employed a cascaded autoencoder to represent video patches. It used a lightweight network to select local patches of interest and then applied a complex 3D convolutional network to detect anomalies. The lightweight network can filter simple normal patches to reduce computational costs and save processing time. S$^2$-VAE \cite{S2-VAE} first detected the foreground and retained only the cell containing the object as input. And then, a shadow generative network was used to fit the data distribution. The output was fed to another deep generative network to model normality. Wu \textit{et al.} \cite{DeepOC} proposed a deep one-class neural network (DeepOC). Specifically, DeepOC used stacked auto-encoders to generate low-dimensional features for frame and optical flow patches and simultaneously trained the OC classifier to make these representations more compact.

Spatial-Temporal Cascade Auto-Encoder (ST-CaAE) \cite{ST-CaAE} first used an adversarial autoencoder to identify anomalous videos and excluded simple regular patches. The retained patches were fed to a convolutional autoencoder, which discriminated anomalies based on reconstruction errors. Liu \textit{et al.} \cite{AST-AE} proposed an Attention augmented Spatial-Temporal Auto-Encoder (AST-AE) that equated frames in spatial dimensions into $8\times 8$ parts and models spatial and temporal information using CNN and LSTM, respectively. In the downstream anomaly detection stage, AST-AE only retained significant regions with large prediction errors to calculate the anomaly score.
\subsection{Object-level Methods}

\begin{table}[t]
  \centering
  \caption{Object-level UVAD Methods.}
  \label{tab:mbj}
    \begin{tabular}{@{}cp{1.5cm}p{1.3cm}p{2.6cm}p{7.2cm}@{}}
  \toprule
  \textbf{Year} & \multicolumn{1}{c}{\textbf{Method}} & \multicolumn{1}{c}{\textbf{Detector}} & \multicolumn{1}{c}{\textbf{Decision Logic}}      & \multicolumn{1}{c}{\textbf{Contributions}}        \\ \midrule
  2017  & LDGK \cite{LDGK}    & Fast R-CNN      & Anomaly score of the detected object proposal & Integrating a generic CNN and environment-related anomaly detector to detect video anomalies and record the cause of the anomalies.        \\ \midrule
  2018  & DCF \cite{DCF}    & YOLO      & Classification       & Extracting foreground objects by object detection models and Kalman filtering and discriminating anomalies by pose and motion classification.       \\ \midrule
  2019  & OC-AE \cite{OC-AE}  & SSD       & One-versus-rest binary classification      & Proposing an object-centric convolutional autoencoder to encode motion and appearance and discriminating anomalies using a one-versus-rest classifier. \\ \midrule
  2021  & Background-Agnostic \cite{Background-Agnostic}  & SSD-FPN, YOLOv3 & Binary classification        & Using a set of autoencoders to extract the appearance and motion features of foreground objects and then using a set of binary classifiers to detect anomalies.  \\ \midrule
  2021  & Multi-task \cite{Multi-task}   & YOLOv3    & Binary classification        & Training a 3D convolutional neural network to generate discriminative representation by performing multiple self-supervised learning tasks.      \\ \midrule
  2021  & OAD \cite{OAD}    & YOLOv3    & Clustering     & An online VAD method with asymptotic bounds on the false alarm rate, providing a  procedure for selecting a proper decision threshold. \\ \midrule
  2021  & HF2-VAD \cite{HF2VAD}      & Cascade R-CNN   & Prediction error    & A hybrid framework that seamlessly integrates sequence reconstruction and frame prediction to handle video anomaly detection.       \\ \midrule
  2020  & VEC \cite{VEC}    & Cascade R-CNN   & Cube construction error    & Proposing a video event completion framework to exploit advanced semantic and temporal contextual information for video anomaly detection.      \\ \midrule
  2022  & BiP \cite{BiP}    & Cascade R-CNN   & Appearance and motion error   & Proposing a bi-directional architecture with three consistency constraints to regularize the prediction task from the pixel, cross pattern, and temporal levels.     \\ \midrule
  2022  & HSNBM \cite{HSNBM}  & Cascade R-CNN   & Frame and object prediction error & Designing a hierarchical scene normative binding modeling framework to detect global and local anomalies.       \\ \bottomrule
\end{tabular}
\end{table}

The emergence of high-performance object detection models \cite{RCNN,FPN,YOLO} provides a new idea for GVAED, i.e., using a pre-trained object detector to extract the objects of interest from the video sequence before normality learning. Compared with the frame-level and patch-level methods, the object-level methods \cite{Background-Agnostic,sun2023hierarchical,liu2023osin,yu2023video} enable the model to ignore redundant background information and focus on modeling the behavioral interactions of foreground objects. In addition to outperforming object-free methods in terms of performance, object-level methods are also considered feasible to investigate scene-adaptive GVAED models. Existing studies \cite{Background-Agnostic,Multi-task} show that object-level methods perform significantly better than other methods on multi-scene datasets such as ShanghaiTech \cite{FFP}. Table~\ref{tab:mbj} compares the object detectors, decision logic, and main contributions of existing object-level methods.

Ryota \textit{et al.} \cite{LDGK} attempted to describe anomalous events in a human-understandable form by detecting and analyzing the classes, behaviors, and attributes of specific objects. The proposed LDGK model first used multi-task learning to obtain anomaly-related semantic information and then inserted an anomaly detector to analyze scene-independent features to detect anomalies. The DCF \cite{DCF} used a pose classifier and an LSTM network to model the spatial and motion information of the detected objects, respectively. \cite{OC-AE} formalizes UVAD as a one-versus-rest binary classification task. The proposed OC-AE first encoded the motion and appearance of selected objects and then clustered the training samples into normal clusters. An object is considered anomalous in the inference stage if the one-versus-rest classifier's highest classification score is negative. Its extension, the Background-Agnostic framework \cite{Background-Agnostic}, introduced instance segmentation, allowing the model to focus only on the primary object. In addition, the authors used pseudo-anomaly examples to perform adversarial learning to improve the appearance and motion auto-encoders.

To make full use of the contextual information, Yu \textit{et al.} \cite{VEC} proposed a Video Event Completion (VEC) method that used appearance and motion as cues to locate regions of interest. VEC recovered the original video events by solving visual completion tests to capture high-level semantics and inferring deleted patches. Georgescu \textit{et al.} \cite{Multi-task} designed several self-supervised learning tasks, including discrimination of forward/backward moving objects, discrimination of objects in continuous/intermittent frames, and reconstruction of object-specific appearance. In the testing phase, anomalous objects would lead to large prediction discrepancies.

Doshi and Yilmaz \cite{OAD} proposed an Online Anomaly Detection (OAD) scheme that used detected object information such as location, category, and size as input to a clustering model to detect anomalous events. HF$^2$-VAD \cite{HF2VAD} seamlessly integrated frames reconstruction and prediction. It used memory to record the normal pattern of optical flow reconstruction and captured the correlation between RGB frames and optical flow using a conditional variation auto-encoder.

Chen \textit{et al.} \cite{BiP} proposed a Bidirectional Prediction (BiP) architecture with three consistency constraints. Specifically, prediction consistency considered the symmetry of motion and appearance in forward and backward prediction. Association consistency considered the correlation between frames and optical flow, and temporal consistency was used to ensure that BiP can generate temporally consistent frames. 

In summary, object-level methods, employing well-trained object detection/segmentation models to isolate significant foreground targets from video frames and developing scene-independent anomaly detection models through analysis of target-specific attributes, offer notable advantages over frame-level and patch-level approaches in real-world cross-scene datasets. However, these methods face challenges in capturing the interaction between scenes and backgrounds, leading to performance degradation in handling scene-specific anomalous events, such as a person walking on a motorway. Addressing this limitation, Liu \textit{et al.} \cite{liu2023osin} explored the semantic interaction between prototypical features of foreground targets and the background scene using memory networks. Alternatively, instance segmentation proves more effective in modeling target-scene interactions. For instance, the Hierarchical Scene Normality-Binding Modeling (HSNBM) framework \cite{HSNBM} attempted to dissect global and local  scenes, which introduced a scene object-binding frame prediction module to capture the relationship between foreground and background through scene segmentation. Looking ahead, object-level methods with object detection or instance segmentation will play a crucial role in discovering anomalous events in real-world highly dynamic environments, such as autonomous driving and intelligent industries.

\section{Weakly-supervised Abnormal Event Detection}~\label{sec4}
Using weakly semantic video-level labels to supervise the model was first proposed by Sultani \textit{et al.} \cite{MIR} in 2018, laying the foundation for WAED based on multiple instance learning \cite{lv2023unbiased,zhang2023exploiting,park2023normality}. The synchronously released UCF-crime dataset collected 13 classes of real-world criminal behaviors and provided video-level labels for training sets. Following researchers \cite{GCLNC,RTFM} made UVAD datasets meet WAED requirements by moving some anomalous test videos to the training set, introducing various WAED benchmarks such as the reorganized UCSD Ped2 \cite{SMR} and ShanghaiTech Weakly \cite{GCLNC,GCLNC+}. In 2020, the XD-Violence \cite{HL-Net} dataset extended GVAED to multimodal signal processing. 

This section is dedicated to providing an in-depth exploration of existing WAED models, with a focus on their categorization into unimodal and multimodal approaches based on the input data modalities. This taxonomy is instrumental in guiding the development of GVAED methods, fostering the transition from video processing to a broader multimodal understanding communities. \textbf{Unimodal models} \cite{MIR,MIST,RTFM,GCLNC,GCLNC+,SRF,STA}, similar to UVAD techniques, primarily rely on RGB frames as input data. However, they distinguish themselves by directly computing the anomaly score. These models center their efforts on analyzing successive RGB frames to detect anomalies within the videos. In contrast, \textbf{multimodal models} \cite{HL-Net+,HL-Net,MACIL-SD,MSAF,chen2023tevad} aim to leverage diverse data sources, including video, audio, text, and optical flow, to extract effective anomaly-related clues. These methods harness the power of multiple modalities to enhance the overall understanding of anomalies, making them more robust and versatile in capturing complex abnormal events. This categorization scheme not only clarifies the distinctions between unimodal and multimodal WAED models but also sets the stage for the evolution of GVAED techniques that integrate various data modalities, paving the way for a more comprehensive approach to anomaly detection and event understanding.

\subsection{Unimodal Models}
   The unimodal WAED model typically slices the unedited video into several fixed-size clips. They consider each clip as an instance, and all clips from a video form a bag with the same video-level label. And then, pre-trained feature extractors, such as Convolutional 3D (C3D) \cite{C3D}, Temporal Segment Networks (TSN) \cite{TSN}, and Inflated 3D (I3D) \cite{I3D}, is used to extract the spatial-temporal features of the examples. Generally, the scoring module takes deep representations as input and calculates the anomaly score for each instance with the supervision of video-level labels. 
  
  The MIL ranking framework \cite{MIR} introduced multiple instance learning to GVAED for the first time, using a 3-layer Fully Connected Network (FCN) to predict high anomaly scores for anomalous clips and introducing sparsity and smooth constraints to avoid drastic fluctuations in the score curve. Zhu and Newsam \cite {MAF} considered motion as the key to WAED performance. To this end, they proposed a temporal augmented network to learn motion-aware features and used attention blocks \cite{liu2023dsdcla} to incorporate temporal context into a MIL ranking model. Snehashis \textit{et al.} \cite{TS-CNN} used a dual-stream CNN to extract spatial and temporal features separately and fed the fused features as spatial-temporal representations into an FCN to perform anomaly classification. The authors compared the performance of different deep CNN architectures (e.g., ResNet- 50 \cite{Resnet}, Inception V3 \cite{Inception}, and VGG-16 \cite{VGG}) for feature extraction.
  
  Zhong \textit{et al.} \cite{GCLNC} treated WAED as a supervised learning task under noisy labels, arguing that the supervised action recognition models can perform anomaly detection after the label noise is removed. In response, they designed a graph convolutional network to correct the labels. \cite{ARNet} proposed Anomaly Regression Network (ARNet) to learn discriminative features WAED. Specifically, ARNet used dynamic multiple-instance learning loss and center loss to enlarge the inter-class distance instances and reduce the intra-class distance of regular instances, respectively.
  
  Waseem \textit{et al.} \cite{CNN-echo} proposed a two-stage WAED framework that first used an echo state network to obtain spatially and temporally aware features. And then, they used a 3D convolutional network to extract spatial-temporal features and fuse them with the features from the first stage as the input to a binary classifier. Tian \textit{et al.} proposed Robust Temporal Feature Magnitude (RTFM) learning by training a feature volume learning function to identify positive examples efficiently. In addition, RTFM utilized self-attention to capture both long and short-time correlations. Muhammad \textit{et al.} \cite{SRF} proposed a self-reasoning framework that uses binary clustering to generate pseudo-labels to supervise the MIL regression models.
  
  The CLustering Assisted Weakly Supervised (CLAWS) learning with normalcy sppression in \cite{CLAWS} proposed a random batch-based training strategy to reduce the correlation between batches. In addition, the authors introduced a loss based on clustering distance to optimize the network to weaken the effect of label noise. Kamoona \textit{et al.} \cite{DTED} proposed a Deep Temporal Encoding-Decoding (DTED) to capture the temporal evolution of videos over time. They treated instances of the same bag as sequential visual data rather than as independent individuals. In addition, DTED uses joint loss to optimize to maximize the average distance between normal and abnormal videos.
  
  The Weakly-supervised Temporal Relationship (WSTR) learning framework \cite{WSTR} enhanced the model's discriminative power by exploring the temporal relationships between clips. The proposed transformer-enabled encoder converts the task-irrelevant representations into task-specific features by mining the semantic correlations and positional relationships between video clips. Weakly Supervised Anomaly Localization (WSAL) \cite{WASL} performed anomaly detection by fusing temporal and spatial contexts and proposed a higher-order context encoding model to measure temporal dynamic changes. In addition, the authors collected a dataset called TAD for traffic anomaly detection.
  
  Feng \textit{et al.} \cite{MIST} proposed a Multi-Instance Self-Training (MIST) framework consisting of a multi-instance pseudo label generator and a self-guided attention-enhancing feature encoder for generating more reliable fragment-level pseudo labels and extracting task-specific representations, respectively. Liu \textit{et al.} \cite{SMR} proposed a Self-guiding Multi-instance Ranking (SMR) framework that used a clustering module to generate pseudo labels to aid the training of supervised multi-instance regression models to explore task-relevant feature representations. The authors compared the performance of different recurrent neural networks in exploring temporal correlation. Spatial-Temporal Attention (STA) \cite{STA} explored the connection between example local representations and global spatial-temporal features through a recurrent cross-attention operation and used mutual cosine loss to encourage the enhanced features to be task specific.

\subsection{Multimodal Models}

\begin{table}[]
  \centering
  \caption{Multimodal WAED Models.}
  \label{tab:dmt}
  \begin{tabular}{@{}cp{1.5cm}p{1.8cm}p{9.1cm}@{}}
  \toprule
  \multicolumn{1}{c}{\textbf{Year}} & \multicolumn{1}{c}{\textbf{Method}} & \multicolumn{1}{c}{\textbf{Input Modality}} & \multicolumn{1}{c}{\textbf{Contributions}}     \\ \midrule
  2020  & HL-Net \cite{HL-Net}  & Video + Audio    & Collecting the XD-Violence violence detection datasets and proposing a three-branch neural network model for multimodal anomaly detection.       \\ \midrule
  2021  & FVAI \cite{FVAI}   & Video + Audio    & Proposing a pooling-based feature fusion strategy to fuse video and audio information to obtain more discriminative feature representations.       \\ \midrule
  2022  & SC \cite{SC}      & Video + Audio    & Proposing an audio-visual scene classification dataset containing 5 classes of anomalous events and a deep classification model.     \\ \midrule
  2022  & MACIL-SD \cite{MACIL-SD}     & Video + Audio    & Proposing a modality-aware contrastive instance learning with a self-distillation strategy to address the modality heterogeneity challenges.       \\ \midrule
  2022  & ACF \cite{ACF}   & Video + Audio    & Proposing a two-stage multimodal information fusion method for violence detection that first refines video-level labels into clip-level labels.        \\ \midrule
  2022  & MSAF \cite{MSAF}   & Video + Audio, Video + Optical flow    & Proposing multimodal labels refinement to refine video-level ground truth into pseudo-clip-level labels and implicitly align multimodal information with multimodal supervise-attention fusion network.        \\ \midrule
  2022  & MD \cite{MD}     & Video + Audio + Flow      & Using mutual distillation to transfer information and proposing a multimodal fusion network to fuse video, audio, and flow features. \\ \midrule
  2022  & HL-Net+ \cite{HL-Net+}      & Video + Audio    & Introducing coarse-grained violent frame and fine-grained violence detection tasks and proposing audio-visual violence detection network.       \\ \midrule
  2022  & AGAN \cite{AGAN}   & Video + Audio    & Using cross-modal interaction to enhance video and audio and computing high-confidence violence scores using temporal convolution. \\ \bottomrule
  \end{tabular}
\end{table}

The emergence of TV shows and streaming media has broadened the application scope of GVAED technicals, transitioning them from traditional offline surveillance video analysis to online video stream detection. Unlike surveillance videos, which typically consist of only RGB images, most online video content, including vlogs, live streams, and talk shows, incorporates multiple modalities such as language, speech, and subtitle text. Extracting anomaly-related cues from these diverse data modalities exceeds the capabilities of current unimodal methods.

Real-world data are heterogeneous, and effectively exploiting the complementary nature of multimodal data is the key to developing robust and efficient GVAED models. Due to the limitation of datasets, most existing works \cite{HL-Net,HL-Net+,MACIL-SD} focused on video and audio information fusion to detect violent behaviors from surveillance videos. Moreover, inspired by the frame-level multi-stream UVAD models \cite{DSTAE,AMAE}, recent work \cite{MSAF} considered RGB frames and optical flow as different modalities. We display the modalities and principles of existing multimodal GVAED models \cite{HL-Net+,HL-Net,SC,FVAI,MACIL-SD,ACF,MSAF,MD,AGAN} in Table~\ref{tab:dmt}.

Wu \textit{et al.} \cite{HL-Net} released the first multimodal GVAED dataset and proposed a three-branch network called HL-Net for multimodal violence detection. Specifically, the similarity branch used a similarity prior to capture long-range correlations. In contrast, the proximity branch used proximity prior to capture local location relationships, and the scoring branch dynamically captured the proximity of predicted scores. Experimental results demonstrated the multimodal data's positive impact on GVAED. The following MACIL-SD in \cite{MACIL-SD} utilized a lightweight dual-stream network to overcome the heterogeneity challenge. It used self-distillation to transfer unimodal visual knowledge to audio-visual models to narrow the semantic gap between multimodal features.

Researchers \cite{FVAI,AGAN} attempted to explore more effective feature extraction and multimodal information fusion strategies. For example, Pang \textit{et al.} \cite{FVAI} proposed to use a bilinear pooling mechanism to fuse visual and audio information and encourage the model to learn from each other to obtain a more effective representation. Audio-Guided Attention Network (AGAN) \cite{AGAN} first used a deep neural network to extract video and audio features and then enhanced the features in the temporal dimension using a cross-modal perceptual local arousal network.

Wei \textit{et al.} \cite{ACF} proposed a two-stage multimodal information fusion method, which first refines video-level hard labels into clip-level soft labels and then uses an attention module for multimodal information fusion. Their extension work, Multimodal Supervised Attentional Augmentation Fusion (MSAF) \cite{MSAF}, used attention fusion to align information and achieved implicit alignment of multimodal data.

Shang \textit{et al.} \cite{MD} observed that existing models are limited by small datasets and proposed mutual distillation to transfer information from large-scale datasets to small datasets. They proposed a multimodal attention fusion strategy to fuse RGB images, audio, and flow to obtain a more discriminative representation. \cite{SC} introduced an audio-visual scene classification task and released a multimodal dataset. The authors try different deep networks and fusion strategies to explore the most effective classification model. 

\section{Supervised Video Anomaly Detection}~\label{sec6}
Supervised video anomaly detection requires frame-level or pixel-level labels to supervise models to distinguish between normal and anomalies. Therefore, it is often considered a classification task rather than a mainstream GVAED scheme. On the one hand, collecting fine-grained labeled anomalous samples is time-consuming. On the other hand, the anomalous behavior occurs gradually, and the degree of anomaly is a relative value, while manual labeling can only provide discrete $0/1$ labels, which cannot adequately describe the severity and temporal continuity of video anomalies. Existing SVAD methods usually consider VAD a binary classification task under data imbalance conditions. However, game engines can simulate various types of anomalous events and provide frame-level and pixel-level personalized annotations. With the penetration of synthetic datasets in vision tasks, supervised training of GVAED models with virtual anomalies is expected to become possible. Researchers need to focus on the domain adaptation problem posed by synthetic datasets, i.e., how to cope with the covariate shifts between synthetic data and the real-world surveillance video and the ensuing performance degradation. Moreover, although the training set contains partially labeled anomalies, SVAD models still need to consider how to reasonably generalize the anomalies to detect unseen anomalous events in real-world scenarios. SVAD is an open-set recognition task rather than a supervised binary classification. 

\section{Fully-Unsupervised Video Anomaly Detection}~\label{sec5}

\begin{figure}[t]
  \centering
  \includegraphics[width=\textwidth]{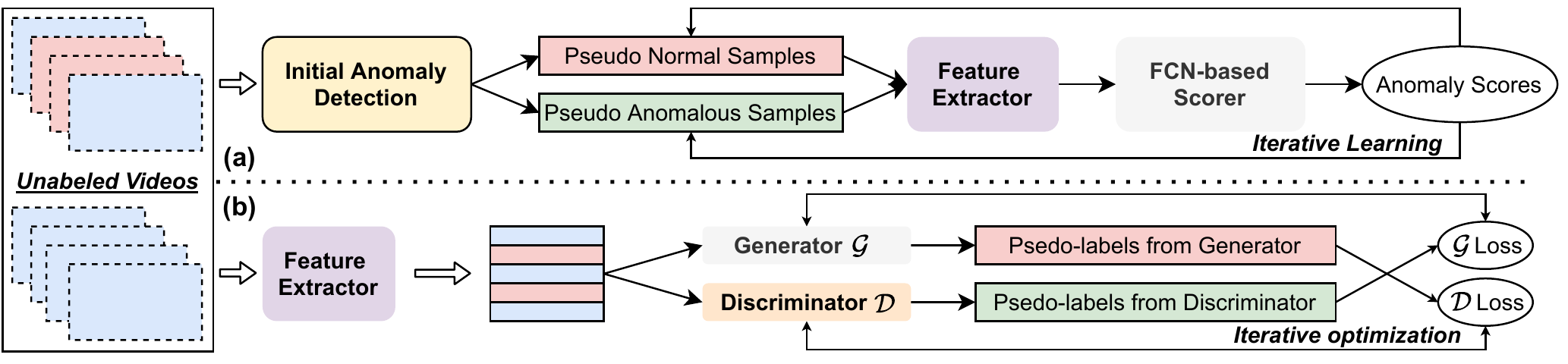}
  \caption{Workflow of two representative FVAD methods: (a) SDOR \cite{SDOR} and (b) GCL \cite{GCL}. Given the unlabeled videos, the SDOR first divided them into pseudo-normal and anomalous sets by initial anomaly detection. GCL introduces cross-supervision to train the generator $\mathcal{G}$ and discriminator $\mathcal{D}$ to learn anomaly detectors. The pseudo-labels from $\mathcal{G}$ and $\mathcal{D}$ are used to compute each other's losses.}
  \label{fig:SDOR}
\end{figure}

Fully-unsupervised Video Anomaly Detection (FVAD) does not limit the composition of the training data and requires no data annotation. In other words, FVAD tries to learn an anomaly detector from the random raw data, which is a newly emerged technical route in recent years.

Ionescu \textit{et al.} \cite{Unmasking} introduced the unmasking technique to computer vision tasks, proposing an FVAD framework that requires no training sequences. They iteratively trained a binary classifier to distinguish two consecutive video sequences and simultaneously removed the most discriminative features at each step. Inspired by \cite{Unmasking}, Liu \textit{et al.} \cite{HS} tried to establish the connection between heuristic unmasking and multiple classifiers two sample tests to improve its testing capability. In this regard, they proposed a history sampling method to increase the testing power as well as to improve the GVAED performance. Li \textit{et al.} \cite{Deep-UAD} first used a distribution clustering framework to identify the possible anomalous samples in the training data, and then used the clustered subset of normal data to train the auto-encoder. An encoder that can describe normality was obtained by repeating normal subset selection and representation learning.

The recent representative FVAD works are Self-trained Deep Ordinal regression (SDOR) \cite{SDOR} and Generative Cooperative Learning (GCL) \cite{GCL}, which attempted to learn anomaly scorers from unlabeled videos in an end-to-end manner, as shown in Fig.~\ref{fig:SDOR}(a) and ~\ref{fig:SDOR}(b). Specifically, SDOR \cite{SDOR} first determined the initial pseudo-normal and abnormal sets and then computed the abnormal scores using pre-trained ResNet-50 and FCN. The representation module and the scorer were optimized iteratively in a self-training manner. Moreover, Lin \textit{et al.} \cite{aaai21} looked at the pseudo label generation process in SDOR from a causal inference perspective and proposed a causal graph to analyze confounding effects and eliminate the impact of noisy pseudo labels. In addition, their proposed CIL model improved significantly by performing counterfactual inference to capture long-range temporal dependencies.

In contrast, GCL \cite{GCL} attempted to exploit the low-frequency nature of anomalous events. It included a generator $\mathcal{G}$ and a discriminator $\mathcal{D}$, which were supervised by each other in a cooperative rather. The generator primarily generated representations for normal events. While for anomaly events, the generator used negative learning techniques to distort the anomaly representation and generated pseudo-labels to train $\mathcal{D}$. The discriminator estimated the probability of anomalies and created pseudo labels to improve $\mathcal{G}$. The scarcity and infrequent occurrence of anomalies provide valuable insights into the development of FVAD. Hu \textit{et al.} \cite{TMAE} leveraged the rarity of anomalies, operating under the assumption that the small number of anomalous samples in the training set has a limited impact on the normality of the model learning process. Inspired by the Masked Auto-Encoder (MAE) \cite{he2022masked}, their proposed TMAE learned representations using a visual transformer performing a complementary task. Notably, MAE \cite{he2022masked} applied masks primarily to 2D images, whereas video anomalies are closely linked to temporal information. To address this challenge, TMAE first identified video foregrounds and constructed temporal cubes to serve as masked objects, ensuring a more comprehensive approach to anomaly detection in video data.

\begin{table}[t]
  \centering
  \caption{EER and AUC Comparison of Early Unsupervised Methods on Benchmark Datasets.}
  \label{tab:wjdxn}
  \resizebox{\textwidth}{!}{
  \begin{tabular}{@{}clccccccc@{}}
  \toprule
  \textbf{Year} & \multicolumn{1}{c}{\textbf{Method}}  & \textbf{Ped1 AUC} & \textbf{Ped1 EER} & \textbf{Ped2 AUC} & \textbf{Ped2 EER} & \textbf{Avenue AUC} & \textbf{Avenue EER} & \textbf{ShanghaiTech AUC} \\ \midrule
  2015 & DRAM \cite{T48}   & 92.1     & 16.0       & 90.8     & 17.0       & -     & -     & -   \\
  2015 & STVP \cite{T41}   & 93.9     & 12.9     & 94.6     & 10.6     & -     & -     & -   \\
  2016 & CMAC \cite{T106}  & 85.0       & -   & 90.0       & -   & -     & -     & -   \\
  2016 & FF-AE \cite{FF-AE} & 81.0       & 27.9     & 90.0      & 21.7     & 70.2       & 25.1       & 60.9     \\
  2017 & DEM \cite{T84}   & 92.5     & 15.1     & -   & -   & -     & -     & -   \\
  2017 & CFS \cite{T108}  & 82.0       & 21.1     & 84.0       & 19.2     & -     & -     & -   \\
  2017 & WTA-AE \cite{T66}   & 91.9     & 15.9     & 92.8     & 11.2     & 82.1       & 24.2       & -   \\
  2017 & EBM \cite{T100}  & 70.3     & 35.4     & 86.4     & 16.5     & 78.8       & 27.2       & -   \\
  2017 & CPE \cite{T77}   & 78.2     & 24.0       & 80.7     & 19.0       & -     & -     & -   \\
  2017 & LDGK \cite{LDGK} & -   & -   & 92.2     & 13.9     & -     & -     & -   \\
  2017 & sRNN \cite{sRNN} & -   & -   & 92.2     & -   & 81.7       & -     & 68.0      \\
  2017 & GANS \cite{T54}   & 97.4     & 8.0        & 93.5     & 14.0       & -     & -     & -   \\
  2017 & OGNG \cite{T67}   & 93.8     & -   & 94.0       & -   & -     & -     & -   \\
  2018 & FFP \cite{FFP}  & 83.1     & -   & 95.4     & -   & 85.1       & -     & 72.8     \\
  2018 & PP-CNN \cite{T45}   & 95.7     & 8.0        & 88.4     & 18.0       & -     & -     & -   \\
  2019 & FAED \cite{T28}   & 93.8     & 14.0       & 95.0       & -   & -     & -     & -   \\
  2019 & NNC \cite{T30}   & -   & -   & -   & -   & 88.9       & -     & -   \\
  2019 & OC-AE \cite{OC-AE} & -   & -   & 97.8     & -   & 90.4       & -     & 84.9     \\
  2019 & AMC \cite{AMC}  & -   & -   & 96.2     & -   & 86.9       & -     & -   \\
  2019 & MLR \cite{T26}   & 82.3     & 23.5     & 99.2     & 2.5      & 71.5       & 36.4       & -   \\
  2019 & memAE \cite{memAE}        & -   & -   & 94.1     & -   & 83.3       & -     & 71.2     \\
  2019 & MLEP \cite{MLEP} & -   & -   & -   & -   & 92.8       & -     & 76.8     \\
  2019 & BMAN \cite{BMAN} & -   & -   & 96.6     & -   & 90.0 & -     & 76.2     \\
  2020 & Street Scene \cite{T6}    & 77.3     & 25.9     & 88.3     & 18.9     & 72.0 & 33.0 & -   \\
  2020 & IPR \cite{T97}   & 82.6     & -   & 96.2     & -   & 83.7       & -     & 73.0 \\
  2020 & DFSN \cite{T27}   & 86.0       & 23.3     & 94.0       & 14.1     & 87.2  & 18.8       & -   \\
 \bottomrule
  \end{tabular}}
\end{table}

\begin{table}[]
  \centering
  \caption{AUC Comparison of Recent Unsupervised and \textit{Fully-unsupervised (Marked in Italics)} Methods on Benchmark Datasets.}
  \label{tab:zxxn}
  \resizebox{\textwidth}{!}{
  \begin{tabular}{@{}clccc|clccc@{}}
  \toprule
  \textbf{Year} & \multicolumn{1}{c}{\textbf{Method}} & \textbf{Ped2} & \textbf{Avenue} & \textbf{ShanghaiTech} & \textbf{Year} & \multicolumn{1}{c}{\textbf{Method}} & \textbf{Ped2} & \textbf{Avenue} & \textbf{ShanghaiTech} \\ \midrule
  2020 & MNAD-R \cite{MNAD}    & 90.2          & 82.8            & 69.8 &  2020 & MNAD-P \cite{MNAD}    & 97.0          & 88.5            & 70.5     \\
  2020 & DD-GAN \cite{DD-GAN}          & 95.6        & 84.9            & 73.7  & 2020 & \textit{SDOR} \cite{SDOR} & 83.2 &- &-    \\
  2020 & ASSAD \cite{ASSAD}           & 97.8        & 86.4            & 71.6     &
  2020 & FSSA \cite{FSSA}    & 96.2        & 85.8            & 77.9  \\
  2020 & VEC \cite{VEC}     & 97.3        & 89.6            & 74.8     &
  2020 & Multispace\cite{TAC-Net}     & 95.4        & 86.8            & 73.6     \\
  2020 & CDD-AE \cite{CDD-AE}          & 96.5        & 86.0      & 73.3     &
  2021 & CDD-AE+ \cite{CDD-AE+}          & 96.7        & 87.1      & 73.7     \\
  2021 & Multi-task (object level) \cite{Multi-task}    & 99.8        & 91.9            & 89.3     &
  2021 & Multi-task (frame level) \cite{Multi-task}     & 92.4        & 86.9            & 83.5     \\
  2021 & Multi-task (late fusion) \cite{Multi-task}     & 99.8        & 92.8            & 90.2     &
  2021 & HF$^2$AVD \cite{HF2VAD}          & 99.3        & 91.1            & 76.2     \\
  2021 & AST-AE \cite{AST-AE}          & 96.6        & 85.2            & 68.8     &
  2021 & ROADMAP\cite{ROADMAP}          & 96.3        & 88.3            & 76.6     \\
  2021 & CT-D2GAN\cite{CT-D2GAN}          & 97.2        & 85.9            & 77.7     &
  2022 & AMAE \cite{AMAE}     & 97.4        & 88.2            & 73.6     \\
  2022 & STM-AE \cite{STM-AE}          & 98.1        & 89.8            & 73.8     &
  2022 & BiP \cite{BiP}      & 97.4        & 86.7            & 73.6     \\
  2022 & AR-AE \cite{AR-AE}      & 98.3        & 90.3            & 78.1     &
  2022 &TAC-Net\cite{TAC-Net}      & 98.1        & 88.8            & 77.2     \\
  2022 & STC-Net \cite{STC-Net}     & 96.7        & 87.8            & 73.1     &
  2022 & HSNBM \cite{HSNBM}           & 95.2        & 91.6            & 76.5     \\ 
  2022 & \textit{CIL(ResNet50)+DCFD} \cite{aaai21} & 97.9& 85.9& - & 2022 & \textit{CIL(ResNet50)+DCFD+CTCE} \cite{aaai21} & 99.4 & 87.3&- \\
  2022 & \textit{CIL(I3D-RGB)+DCFD+CTCE} \cite{aaai21} & 98.7& 90.3& - & 2022 & \textit{GCL$_{PT}$(RESNEXT)} \cite{GCL} & - & - & 78.93 \\
\bottomrule
  \end{tabular}}
\end{table}

\begin{table}[]
  \centering
  \caption{Quantitative Performance Comparison of Weakly-supervised Methods on Public Datasets.}
  \label{tab:rjdxn}
  \resizebox{\textwidth}{!}{
  \begin{tabular}{@{}llcccc@{}}
  \toprule
  \textbf{Method} & \textbf{Feature} & \textbf{UCF-Crime AUC} & \textbf{UCF-Crime FAR} & \textbf{ShanghaiTech AUC} & \textbf{ShanghaiTech FAR} \\ \midrule
  MIR \cite{MIR}     & C3D$^{RGB}$   & 75.40     & 1.90    & 86.30    & 0.15        \\ \midrule
  TCN \cite{TCN}   & C3D$^{RGB}$   & 78.70     &  -& 82.50        & 0.10 \\ \midrule
  Zhong \cite{GCLNC}   & C3D$^{RGB}$    & 80.67    & 3.30      & 76.44       & -        \\ \midrule
  ARNet \cite{ARNet}  & C3D$^{RGB}$    & -     & -     & 85.01       & 0.57        \\
    & I3D$^{RGB}$    & -     & -     & 85.38       & 0.27        \\
    & I3D$^{RGB+Optical\ Flow}$       & -     & -     & 91.24       & 0.10 \\ \midrule
  MIST \cite{MIST}   & C3D$^{RGB}$    & 81.40     & 2.19     & 93.13       & 1.71        \\
    & I3D$^{RGB}$   & 82.30     & 0.13     & 94.83       & 0.05        \\ \midrule
  RTFM \cite{RTFM}   & C3D$^{RGB}$    & 83.28    & -     & 91.51       & -        \\
    & I3D$^{RGB}$   & 84.30     & -     & 97.21       & -        \\ \midrule
  SMR \cite{SMR}    & I3D$^{RGB+Optical\ Flow}$    & 81.70     & -     & -        & -        \\ \midrule
  DTED \cite{DTED}   & C3D$^{RGB}$    & 79.49    & 0.50      & 87.42       & -        \\ \bottomrule
  \end{tabular}}
  \end{table}

\section{Performance Comparison}~\label{sec7}
We collect the performance of existing works on publicly available datasets \cite{T2,T3,FFP,MIR,GCLNC} to quantitatively compare the superiority and present the GVAED development progress. Table~\ref{tab:wjdxn} presents the frame-level AUC and EER of the early UVAD models on UCSD Ped1 \& Ped2 \cite{T2}, and CUHK Avenue \cite{T3} datasets and the frame-level AUC on the ShanghaiTech \cite{FFP} dataset. 
Since the recent UVAD \cite{memAE,STM-AE,AMAE,CDD-AE} and FVAD \cite{SDOR,aaai21,GCL} only report frame-level AUC as the main evaluation metric, we have collated these methods separately in Table~\ref{tab:zxxn}.
The ShanghaiTech dataset was proposed in 2018 with the FFP \cite{FFP} model, so methods before this time were usually tested without this dataset. With the advantage of its data size and quality, ShanghaiTech has become the most widely used UVAD benchmark. An interesting phenomenon is that the object-level methods outperform other frame-level and patch-level models on the cross-scene ShanghaiTech dataset. For example, the frame-level AUC of the Multi-task \cite{Multi-task} model is as high as 90.2\%, which is 12.1\% higher than the state-of-the-art frame-level methods \cite{AR-AE}. It shows that for cross-scene GVAED tasks, using an object detector to separate the foreground object of interest from the scene can effectively avoid interference of the background. In addition, the multi-stream model learns normality in both temporal and spatial dimensions and generally outperforms the single-stream model. The usage frequency shows that UCSD Ped2 \cite{T2}, CUHK Avenue \cite{T3}, and ShanghaiTech \cite{FFP} have become the prevailing benchmarks for UVAD evaluation. Future work should consider testing and comparing the proposed methods on these three datasets.

Table~\ref{tab:rjdxn} presents the performance of WAED methods on the UCF-Cirme \cite{MIR} and ShanghaiTech Weaky \cite{GCLNC} datasets. As mentioned previously, WAED models usually rely on pre-trained feature extractors \cite{C3D,TSN,I3D} to obtain feature representations. Commonly used features include C3D$^{RGB}$, I3D$^{RGB}$, and I3D$^{RGB+Optical\ flow}$. The performance gap of the same model using different features show that the effectiveness of the WAED model is related to the pre-trained feature extractors, with the I3D outperforming the simple $3\times 3\times 3$ convolution-based C3D network due to the separate consideration of temporal information variation. Future WAED work should test the performance of the proposed model on current commonly used features or provide the performance of existing works on emerging features to demonstrate that the performance gain comes from the model design rather than benefiting from a more robust feature extraction network. In addition to detection performance, other metrics are processing speed and deployment cost. 
GVAED typically employs the Average Inference Speed (AIS) as a visual metric to gauge the overhead cost of the model. Figures reported in existing literature often lack direct comparability due to variations in experimental environments and computational platforms. Moreover, Recent advancements in GVAED research, such as object-level methods and weakly-supervised schemes, typically involve intricate data preprocessing and calls to pre-trained models, such as foreground object detection, optical flow estimation, and spatial-temporal feature extraction with well-trained 3D convolutional networks. It remains unclear whether the computational cost and processing time associated with these aspects are factored into the overhead cost of the proposed model. Consequently, reporting inference speed and comparing computational cost are not widespread practice in GVAED research. The few papers providing such results often lack a detailed description of the experimental setup. Nevertheless, we diligently collected the AIS of existing works to offer an intuitive demonstration of the trend toward lighter-weight GVAED research. Acknowledging the impact of image resolution on model inference speed, we follow \cite{ramachandra2020survey} to summarize these data while simultaneously documenting the dataset used for model testing. The results are publicly accessible in our GitHub repository\footnotemark[1] and will be continuously updated.

\section{Challenges and Trends}~\label{sec8}
\subsection{Research Challenges}

\subsubsection{Mock anomalies vs. Real anomalies: How to bridge domain offsets between mock and real anomalies?}
GVAED aims to automatically detect anomalous events in the living environment to provide a safe space for humans. However, the difficulty of collecting anomalies makes most of the existing datasets formulate abnormal events by human simulation, such as the UMN \cite{UMN}, CUHK Avenue \cite{T3}, and ShanghaiTech \cite{FFP}. The mock anomalies are simpler, and their spatial-temporal patterns differ significantly from normal events, resulting in well-trained models difficult to detect complex anomalies. In addition, the set of limited categories of anomalous events conflicts with the diverse nature of real anomalies. As a result, models learned on such datasets perform poorly in real-world scenarios. Therefore, collecting datasets containing various real anomalies and designing models to bridge the gap between mock and real anomalies is an essential challenge for GVAED development.

\subsubsection{Single-scene vs. Multi-scenes: How to develop cross-scenario GVAED models for the real world?}
Mainstream unsupervised datasets \cite{T2,T3} and UVAD methods \cite{MNAD,FFP,CDD-AE} only consider single-scene videos, while the real world always contains multiple scenes, which constitutes another challenge for UVAD methods. Although the UMN \cite{UMN} and ShanghaiTech \cite{FFP} datasets include multiple scenes, the anomalous events of the former are all crowd dispersal, while the 13 scenes of the latter are similar. Therefore, most UVAD methods \cite{FFP,STM-AE} do not consider the scene differences but learn normality directly from the original video as in other single-scene datasets \cite{T2,T3}. Recent researchers \cite{Background-Agnostic,OC-AE} believe that object-level methods are a feasible way to learn scene-invariant normality by extracting specific objects from the scenes and then analyzing the spatial-temporal patterns of objects and backgrounds separately. The multi-scene problem is inescapable for model deployment as it is almost impossible to train a scene-specific model for each terminal device. Developing cross-scene GVAED models using domain adaptation/generalization techniques \cite{wilson2020survey,liu2023distributional,xiang2023two} to learn scene-invariant normality is a definite challenge.

\subsubsection{Real data vs. Synthetic data: How to develop large fine-grained GVAED models using synthetic data?}
Due to the rarity and diversity of anomalies, collecting and labeling anomalous events is time-consuming and laborious. Therefore, researchers \cite{UBnormal} have considered using game engines \cite{engine1,engine2} to synthesize anomaly data. We remain optimistic about this attempt and believe it may lead to new research opportunities for GVAED. While anomaly detection tasks suffer from a lack of data and missing labels. Synthetic data can generate various anomalous samples and provide precise frame-level or even pixel-level annotations, making it possible to develop SVAD models and save data preparation costs for large-scale GVAED model training. However, a concomitant challenge is that covariate shifts between synthetic and real data may make the trained GVAED models not work in real scenes.

\subsubsection{Unimodal vs. Multimodal: How to effectively fuse multimodal data to mine anomaly-related clues?}
Researchers \cite{HL-Net,HL-Net+} are aware of the positive impact of multimodal data (e.g., audio) for GVAED. However, existing works are stuck on the lack of datasets and the validity of model structures. XD-Violence \cite{HL-Net} is the only mainstream multimodal GVAED dataset, but it only contains video and audio, and much data is collected from movies and games rather than the real world. With the popularity of IoT, using various sensors to collect environmental information (e.g., temperature, brightness, and humidity) can assist cameras in detecting abnormal events. However, mining useful clues from valid data and developing efficient multimodal GVAED models need further research, such as task-specific feature extraction from heterogeneous data, semantic alignment of different modalities, anomaly-related multimodal information fusion, amd domain offset bridging in emerging cross-modal GVAED research.

\subsubsection{Single-view vs. Multi-view: How to integrate complementary information from multi-view data?}
In places such as traffic intersections and parks, the same area is usually covered by multiple camera views, deriving another task: anomalous event detection in multi-view videos. Multi-view data can provide more comprehensive environmental awareness data, which is wildly used for re-identification \cite{lin2019multi,yin2021multi}, tracking \cite{tang2018joint} and gaze estimation \cite{lian2018multiview}. However, existing datasets \cite{UMN,T2,T3,FFP} are all single-view, making multi-view GVAED research still a gap. The simplest idea is to combine data from all views to train the same model and determine anomalies through a voting or winner-take-all strategy. However, such approaches are training-costly and do not consider the differences and complementarities between multi-view data. Therefore, multi-view GVAED remains to be investigated.

\subsubsection{Offline vs. Online Detection: How to develop light-weight end-to-end online GVAED models?}
The deployable Intelligent video surveillance systems (IVSS) need to process continuously generated video streams 24/7 online and respond to anomalous events in real-time so that noteworthy clips can be saved in time to reduce storage and transmission costs. Unfortunately, existing GVAED models are designed for public datasets rather than real-time video streams, primarily pursuing detection performance while avoiding the online detection challenges. For example, the dominant prediction-based methods \cite{FFP,MNAD,STM-AE,AMAE} in UVAD route can only give the prediction error of the current input in a single-step execution, while the Informative anomaly score needs are obtained after performing the maximum-minimum normalization over the prediction error of all frames. Although the model can directly determine the current frame as an anomaly with a pre-set error threshold, existing attempts show that the manually selected threshold is unreliable. WAED \cite{MIR,SMR,MIST,RTFM,WASL} can directly output anomaly scores for segments. However, the input to the scoring module is a discriminative spatial-temporal representation rather than the original video. The representations usually rely on pre-trained 3D convolution-based feature extractors \cite{C3D,I3D,TSN}. The time cost is unacceptable for resource-limited terminal devices \cite{ju2023high}. Therefore, developing online detection models is the primary challenge for GVAED deployment, determining its application potential in IVSS and streaming media platforms.

\subsection{Development Trends}

\subsubsection{Data level: Toward real-world GVAED model development for multi-view cross-scene multimodal data}
From single-scene \cite{T2,T3} to multi-scene \cite{FFP,T6}, from real-word vides \cite{UMN} to synthetic data \cite{UBnormal}, and from unimodal \cite{MIR} to multimodal \cite{HL-Net}, GVAED datasets are moving towards large-scale and realistic scenarios. We see this as a positive trend that will continue with the growth of online video platforms and tools. On the one hand, real-world scenarios and anomalies are diverse, so efficient models for real-world applications need to be trained on large-scale datasets that contain various anomalous events. On the other hand, the Internet has made it possible to collect multi-scene and multi-view videos, including sufficient rare anomalous behaviors such as violence and crime. Furthermore, multimodal and synthetic data will be increasingly important in GVAED research. The XD-Violence \cite{HL-Net} dataset has demonstrated the positive impact of multimodal data on GVAED. In the future, with streaming media (e.g., TikTok, Netflix, and Hulu) and online video sites (e.g., YouTube, iQIYI, and Youku), more modal data can be collected. Besides, virtual game engines (e.g., Airsim \cite{engine1} and Carla \cite{engine2}) can synthesize rare anomalous events and provide fine-grained annotations on demand.
The connection of GVAED with other tasks (e.g., multimodal analysis \cite{MSAF} and few-shot learning \cite{FSSA}) will tend to be close, with the latter inspiring the design of GVAED models under specific data conditions.

\subsubsection{Representation level: Adaptive transfer of emerging representation learning and feature extraction means}
Deep learning has enabled spatial-temporal representations to be derived directly from the raw videos in an end-to-end manner without a human prior \cite{wang2023adversarial}. The earlier deep GVAED models benefit from CNNs and pursue complex deep networks to extract more abstract features. For example, the UVAD models attempt to introduce dual-stream networks to learn spatial and temporal representations \cite{CDD-AE,AMAE,STM-AE}, and use 3D convolutional networks to model temporal features \cite{STAE,memAE}. From C3D \cite{C3D} to I3D \cite{I3D}, the WAED models \cite{MIST,RTFM,SMR} benefit from more powerful pre-trained feature extractors and achieves general performance gains on existing datasets \cite{MIR,GCLNC}. We observe that the representation means of WAED will become increasingly sophisticated. New visual representation learning models such as Transformer \cite{li2022self,transformer1,transformer2,huang2022weakly} will drive WAED development. In contrast, UVAD does not pursue abstract representations. Overly powerful deep networks may lead to missing anomalous events as normal due to overgeneralization \cite{memAE,MNAD}. Future researchers should consider using clever representation strategies (e.g., causal representation learning \cite{liu2023learning}) to balance the model's powerful representation of normal events and the limited generalization of abnormal events \cite{amp}. Powerful generative models such as graph learning \cite{graph1,graph2,graph3} and diffusion models \cite{croitoru2022diffusion} are expected to provide more effective normality learning tools for UVAD. In addition, researchers should consider introducing emerging techniques (e.g., domain adaptation \cite{Background-Agnostic,wang2022generalizing} and knowledge distillation \cite{wang2023sampling}) to develop GVAED models for learning scene-invariant representation from multi-scene and multi-view videos.

\subsubsection{Deployment level: Lightweight easy-to-deploy model development for resource-constrained end devices}
Model deployment is an inevitable trend for GVAED development. As mentioned above, the multi-scene and the diversity of anomalies in real-world videos pose new challenges for model design and training, such as online detection, lightweight models, and high view robustness. On the one hand, the computational resources of terminal devices are limited. Most deep GVAED methods are overly pursuing performance at the expense of running costs. On the other hand, existing models are trained offline, which cannot perform real-time detection. Model compression \cite{deng2020model} and knowledge distillation \cite{gou2021knowledge} can drive the development of lightweight GVAED models. Online evolutive learning \cite{li2022mutual,li2022multi}, dge-cloud collaboration, and integrated sensing and control technologies \cite{qian2023bidirectional} enable models to dynamically optimize learnable parameters in complex environments such as modern industry \cite{amp} and intelligent transportation systems \cite{yang2023aide}.

\subsubsection{Methodology level: High-efficiency \& robust GVAED development by integrating different research pathways}
This survey compares the four main GVAED technical routes: UVAD, WAED, SVAD, and FVAD. UVAD has been regarded as the mainstream solution, although WAED gradually dominates in recent years. However, the trend of UVAD is unclear due to its performance saturation on limited datasets \cite{T2}. In addition, the setting of anomalies in UVAD datasets makes UVAD models challenging to work in complex scenes. Self-supervised visual representation technicals (e.g., contrast learning \cite{TAC-Net,contrast1,contrast2,contrast3} and deep clustering \cite{cluster1,cluster2}) may provide new ideas for UVAD. In contrast, WAED has been widely noticed as a research hotspot due to its excellent performance in crime detection \cite{MIR}. In addition, the multimodal video anomaly detection tasks also follow WAED routes. SVAD is once abandoned due to the lack of labels and anomalies. However, it may face new research opportunities with the emergence of synthetic datasets \cite{UBnormal}. In contrast, FVAD can learn directly from raw video data without the cost of training data filtering and annotations, making it a hot research topic. The various routes are not completely independent, and existing works \cite{wu2022self,liu2022collaborative} have started to combine the assumptions of different methods to develop more efficient GVAED models.

\section{Conclusion}~\label{sec9}

This survey is the first to integrate the deep learning-driven technical routes based on different assumptions and learning frameworks into a unified generalized video anomaly event detection framework. We provide a hierarchical GVAED taxonomy that systematically organizes the existing literature by supervision, input data, and network structure, focusing on the recent advances such as weakly-supervised, fully-unsupervised, and multimodal methods. To provide a comprehensive survey of the extant work, we collect benchmark datasets and available codes, sort out the development lines of various methods, and perform performance comparisons and strengths analysis. This survey helps clarify the connections among deep GVAED routes and advance community development. In addition, we analyze research challenges and future trends in the context of deep learning technology development and possible problems faced by GAED model deployment, which can serve as a guide for future researchers and engineers.

\begin{acks}
  This work is supported in part by the China Mobile Research Fund of the Chinese Ministry of Education under Grant No. KEH2310029, the National Natural Science Foundation of China under Grant No. 62250410368, and the Specific Research Fund of the Innovation Platform for Academicians of Hainan Province under Grant No. YSPTZX202314. Additional support is acknowledged from the Shanghai Key Research Laboratory of NSAI and the Joint Laboratory on Networked AI Edge Computing Fudan University-Changan. The work of Yang Liu was financially supported in part by the China Scholarship Council (File No. 202306100221). The authors extend their appreciation to the anonymous reviewers for their valuable comments and suggestions, as well as to the authors of the reviewed papers for their contributions to the field.
\end{acks}

\bibliographystyle{ACM-Reference-Format}
\bibliography{refs}










\end{document}